\documentclass[journal,twoside,web]{ieeecolor}

\usepackage{etoolbox}
\makeatletter
\@ifundefined{color@begingroup}%
{\let\color@begingroup\relax
	\let\color@endgroup\relax}{}%
\def\fix@ieeecolor@hbox#1{%
	\hbox{\color@begingroup#1\color@endgroup}}
\patchcmd\@makecaption{\hbox}{\fix@ieeecolor@hbox}{}{\FAILED}
\patchcmd\@makecaption{\hbox}{\fix@ieeecolor@hbox}{}{\FAILED}

\usepackage{jsen}
\usepackage{cite}
\usepackage{amsmath,amssymb,amsfonts}
\usepackage{algorithm}
\usepackage{graphicx}
\usepackage{textcomp}
\usepackage{wrapfig}

\usepackage{amsmath}
\usepackage{amsfonts}
\usepackage{amssymb}

\usepackage{multirow}
\usepackage{algorithm}

\usepackage{algpseudocode}

\usepackage{array}

\usepackage{booktabs}
\usepackage{cite}
\usepackage{makecell}
\usepackage{setspace}
\usepackage{booktabs}
\usepackage{color}
\usepackage{colortbl}
\usepackage{scalerel}

\def\BibTeX{{\rm B\kern-.05em{\sc i\kern-.025em b}\kern-.08em
    T\kern-.1667em\lower.7ex\hbox{E}\kern-.125emX}}
\definecolor{abstractbg}{rgb}{0.89804,0.94510,0.83137}
\begin{document}
\title{HDA-LVIO: A High-Precision LiDAR-Visual-Inertial Odometry in Urban Environments with Hybrid Data Association}
\author{Jian~Shi, Wei~Wang,~\IEEEmembership{Senior Member,~IEEE}, Mingyang~Qi, Xin~Li, and Ye~Yan
\thanks{Jian Shi, was with  the College of Intelligent Systems Science and Engineering, Harbin Engineering University,  Harbin, 150001, China, (e-mail: bznh-sj@hrbeu.edu.cn)}
\thanks{Wei Wang, was with  the College of Intelligent Systems Science and Engineering, Harbin Engineering University,  Harbin, 150001, China, (e-mail: wangwei407@hrbeu.edu.cn)}
\thanks{Mingyang Qi, was with  the College of Intelligent Systems Science and Engineering, Harbin Engineering University,  Harbin, 150001, China, (e-mail: qimingyang@hrbeu.edu.cn)}
\thanks{Xin Li, was with  the College of Intelligent Systems Science and Engineering, Harbin Engineering University,  Harbin, 150001, China, (e-mail: xinxin\_forever@126.com)}
\thanks{Ye Yan is with the National Innovation Institute of Defense Technology, Academy of Military Sciences China, Beijing
100071, China, and also with the Tianjin Artificial Intelligence Innovation
Center (TAIIC), Tianjin 300450, China, (e-mails: yynudt@126.com).}
\thanks{Wei Wang is the corresponding author.}
}

\IEEEtitleabstractindextext{%
\fcolorbox{abstractbg}{abstractbg}{%
\begin{minipage}{\textwidth}%
\begin{wrapfigure}[19]{r}{3.8in}%
\includegraphics[width=3.7in]{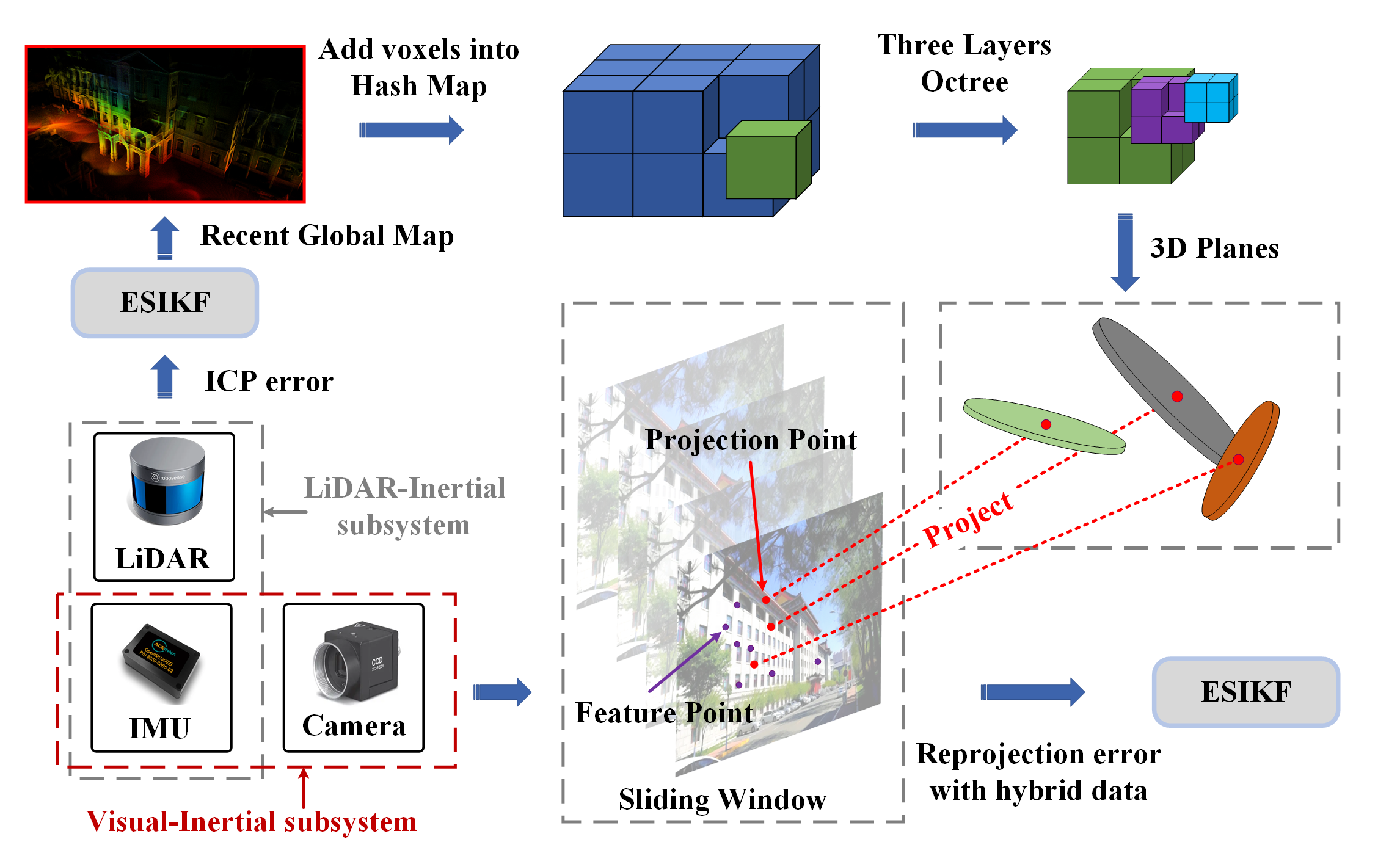}%
\end{wrapfigure}%
\begin{abstract}
To enhance localization accuracy in urban environments, an innovative LiDAR-Visual-Inertial odometry, named HDA-LVIO, is proposed by employing hybrid data association. The proposed HDA-LVIO system can be divided into two subsystems: the LiDAR-Inertial subsystem (LIS) and the Visual-Inertial subsystem (VIS). In the LIS, the LiDAR pointcloud is utilized to calculate the Iterative Closest Point (ICP) error, serving as the measurement value of Error State Iterated  Kalman Filter (ESIKF) to construct the global map. In the VIS, an incremental method is firstly employed to adaptively extract planes from the global map. And the centroids of these planes are projected onto the image to obtain projection points. Then, feature points are extracted from the image and tracked along with projection points using Lucas–Kanade (LK) optical flow. Next, leveraging the vehicle states from previous intervals, sliding window optimization is performed to estimate the depth of feature points. Concurrently, a method based on epipolar geometric constraints is proposed to address tracking failures for feature points, which can improve the accuracy of depth estimation for feature points by ensuring sufficient parallax within the sliding window. Subsequently, the feature points and projection points are hybridly associated to construct reprojection error, serving as the measurement value of ESIKF to estimate vehicle states. Finally, the localization accuracy of the proposed HDA-LVIO is validated using public datasets and data from our equipment. The results demonstrate that the proposed algorithm achieves obviously improvement in  localization accuracy compared to various existing algorithms.
\end{abstract}

\end{minipage}}}

\maketitle

\section{Introduction}
\label{sec:introduction}
\IEEEPARstart{A}{}ccurate localization in urban environments is the foundation for safe operation of autonomous vehicle \cite{kuutti2018survey}. Simultaneous Localization and Mapping (SLAM), capable of achieving vehicle localization and environmental mapping concurrently, is widely utilized for localization in urban environment \cite{cadena2016past}. Depending on the sensors employed, existing SLAM methodologies can be categorized as either LiDAR-based method \cite{zhang2014loam, ASL-SLAM, WangZhoubo} or Visual-based method \cite{qin2018vins, SunTian}. Nevertheless, owing to the sparse nature of LiDAR pointcloud, the accuracy of LiDAR-based SLAM may decrease  in scenes lacking geometric textures, such as tunnels and highway \cite{Chou}. Additionally, LiDAR-based SLAM manifests significant height drift when traversing large-scale movement \cite{shan2018lego}.  Visual-based SLAM methods are susceptible in the environments, such as variations in lighting or absence the color texture, leading to decreased  accuracy in localization \cite{park2017illumination}. Moreover, Monocular Visual based odometry may experience localization drift due to scale uncertainty \cite{strasdat2010scale}. Hence, the SLAM algorithms that exclusively rely on a single sensor cannot achieve high-precision localization within complex urban environments.

In recent years, many researchers have explored the fusion of LiDAR and Visual to improve the localization accuracy. The fusion methods can be classified as LiDAR unenhanced methods \cite{wang2019robust, zuo2019lic, zuo2020lic, zhao2021super, jia2021lvio, gao2022vido} and LiDAR enhanced methods \cite{huang2018joint,zhang2017real,zhang2018laser,shan2021lvi,wang2022mvil, shu2022multi ,tang2023vins,lowe2018complementary,zhu2021camvox,yin2022novel,wang2021vanishing,giubilato2018scale,graeter2018limo,shin2018direct, shin2020dvl,huang2019accurate,huang2020lidar,zheng2022fast,lin2022r,yuan2023sdv}. The LiDAR unenhanced methods  implement LiDAR odometry and Visual odometry independently, and then probabilistically fuses the odometry results to obtain a fused state. In \cite{wang2019robust}, the state estimation results from Visual-Inertial odometry are used to provide motion priors for LiDAR scan registration, thereby improving the localization accuracy of LiDAR odometry in structure-less environments. In \cite{zuo2019lic,zuo2020lic}, an efficient framework based on Multi-State Constraint Kalman Filter (MSCKF) is employed to fuse multi model LiDAR features and Visual features. This method can achieve robust localization in urban environments by utilizing rich environmental features. In \cite{zhao2021super},  LiDAR odometry and Visual odometry are built separately, and the Inertial Measurement Unit (IMU) is utilized to propagate the states between the two odometry systems. This method achieves robust localization in perceptually-degraded environments. 
Additionally,  researchers employ adaptive strategies within the fusion algorithm to enhance the stability of the fusion system.
In \cite{jia2021lvio}, a factor graph is utilized to fuse data from LiDAR, Visual, and IMU. And then an actor-critic method with reinforcement learning is adopted to adaptively adjust the fusion weight of sensors. In \cite{gao2022vido}, a covariance intersection filter is adopted to adaptively estimate the fused state with the uncertainties of the estimation results in LiDAR-Inertial odometry and Visual-Inertial odometry.  
However, since the LiDAR unenhanced methods mentioned above \cite{wang2019robust,zuo2019lic,zuo2020lic,zhao2021super,jia2021lvio,gao2022vido}  cannot fully exploit the complementary characteristics between LiDAR and Visual, these methods will only enhance system stability and not effectively improve the accuracy of the fusion system. 

\begin{figure}[htbp]
	\centering
	\includegraphics[width=2.0in]{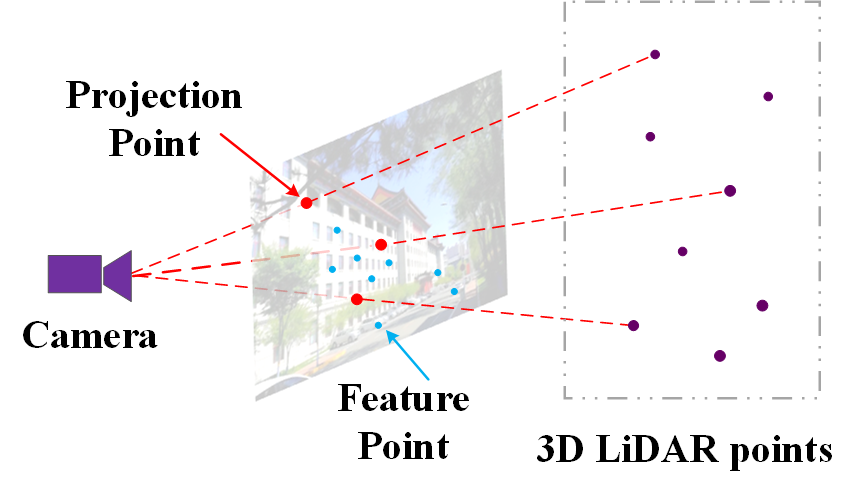}
	\caption{{The projection points and feature points utilized in this paper.} }
	\label{projection_pt}
\end{figure}

In contrast, the LiDAR enhanced method allows the utilization of accurate range information from LiDAR to recover the depth of Visual pixel, thereby enhancing the accuracy of the fusion system.  
At present, the main focus of LiDAR enhanced methods lie in exploring effective ways to associate the accurate range data from LiDAR  and Visual pixel. The association methods can be divided into feature-based methods \cite{huang2018joint,zhang2017real,zhang2018laser,shan2021lvi,wang2022mvil, shu2022multi ,tang2023vins,lowe2018complementary,zhu2021camvox,yin2022novel,wang2021vanishing,giubilato2018scale,graeter2018limo} and direct methods \cite{shin2018direct, shin2020dvl,huang2019accurate,huang2020lidar,zheng2022fast,lin2022r,yuan2023sdv}.  Feature-based methods adopt feature points, such as Harris points \cite{harris1988combined}, to establish associations between LiDAR and Visual. In contrast, direct methods directly project LiDAR points onto images to obtain projection points, which are then utilized to calculate the vehicle states. Feature points and projection points are illustrated in Fig. \ref{projection_pt}, where feature point is represented with blue dot and projection point is represented with red dot.

Feature-based methods utilize the accurate range data from LiDAR to recover depth of feature points. In \cite{huang2018joint}, the 5-DoF state information is obtained through feature-based visual odometry. The scale information is recovered by matching the LiDAR pointcloud with the scale-ambiguous feature map output from the visual odometry using a 1-DoF ICP algorithm. 
In \cite{zhang2017real,zhang2018laser,shan2021lvi,wang2022mvil}, the map points constructed with LiDAR pointcloud and the feature points are simultaneously projected onto the unit sphere of the camera coordinate. And then the kd-tree is utilized to search for three LiDAR points near the feature points. The depth of the feature points is determined by the plane formed by these three LiDAR points. Building upon this kind of approach, in \cite{shu2022multi}, probabilistic description of the plane fitting are  utilized to improve the depth recover accuracy of feature points. Concurrently, the delayed triangulation is  employed to recover the depths of image feature points that lack associated with LiDAR points. However, since the three points selected in this method may not satisfy the plane assumption, the associated depth exists significant deviation. Therefore, in \cite{tang2023vins}, five points are chosen and plane verification is performed to ensure effective association with image feature points and to guarantee the accuracy of depth association. 
In cases of sparse LiDAR point distribution, some feature points may be situated at a considerable distance from the five nearest LiDAR points, result in challenges for accurate data association. 

Furthermore, in \cite{lowe2018complementary}, the depth of feature points is estimated by ray casting onto the surfel map constructed from LiDAR points. In \cite{zhu2021camvox, yin2022novel}, the map constructed from LiDAR is transformed into a depth map consistent with the image dimensions and projected onto the image to form an RGB-D image. This RGB-D image is then utilized to establish data association by identifying the nearest projected point for each image feature point. In \cite{wang2021vanishing}, the LiDAR pointcloud is stored in the voxel grid and plane fitting is performed on the LiDAR points within the voxel. Image feature points are associated with planes in voxels through ray projection.
In \cite{giubilato2018scale}, under the assumption of locally planar environment, LiDAR points are firstly projected onto the image and the depth of feature points within a certain range around the projection points is initialized. And further depth optimization is performed based on iSAM2 \cite{kaess2012isam2}. In \cite{graeter2018limo}, the LiDAR pointcloud of the current frame is fully projected onto the image plane, and LiDAR projection points near the image feature points are extracted for plane fitting. The depth of visual features is obtained by intersecting the Visual rays corresponding to the features with the fitted planes. 
However, due to the sparse nature of LiDAR pointcloud, effectively associating LiDAR points with image feature points will be challenging.

For the direct method, 3D LiDAR points are projected onto the image to derive projection points, achieving data association between LiDAR and Visual.
In \cite{shin2018direct, shin2020dvl}, the projection points in adjacent image frames are associated through optical flow tracking of the projection points, which are then employed to construct reprojection error for the Visual odometry. Furthermore, beyond utilizing LiDAR points directly, researchers have also explored techniques for extracting diverse features from LiDAR pointcloud to derive projection points. In \cite{huang2019accurate}, environment planes are extracted from the LiDAR pointcloud, and corresponding plane patches are extracted from the image. Both the plane patches and projection points are used together for state estimation.  In \cite{huang2020lidar}, point and line features in the environment are extracted from the LiDAR pointcloud  to obtain projection points. Nonetheless, owing to the sparse nature of LiDAR pointcloud, it may pose challenges in effectively extracting feature points to acquire projection points. 
In addition to selection data from the LiDAR pointcloud to obtain projection points, in \cite{lin2022r}, points selected from the global map in the current image viewpoint are projected into the image, and the projection points are used for optical flow tracking and state estimation by minimizing the reprojection error. The estimated results are then used as initial values for updating the state through minimizing photometric errors. In \cite{yuan2023sdv}, LiDAR projection points with large photometric gradients are utilized for optical flow tracking and state estimation. However, optical flow tracking may suffer errors when dealing with projection points that lack obviously illumination intensity gradients. Furthermore, state errors used for projection may introduce biases in the depth correlation of these projection points.


To enhance the accuracy of the LiDAR-Visual-Inertial odometry, we propose a novel hybrid data association method to fuse LiDAR and Visual data in this work. The primary contributions of this article are summarized as follows:
\begin{itemize}
	\item Utilizing the continuity of the environment plane, a novel method is proposed for selecting projection points to mitigate significant errors resulting from pose inaccuracies. Initially, the recent global map is voxelized and organized using a hash table for efficient plane management. Subsequently, an octree map is employed to adaptively extract planes from the map points within each voxel. Next, an incremental plane extraction method is proposed to ensure rapid update of the environment plane when changes occur in recent global map. Finally, the center point of the extracted plane is projected onto the image to obtain projection points, thereby achieving stable depth associations for the projection points.
	\item A robust  depth estimation method is presented for feature points. Firstly, we utilize the pose output from VIS over a time interval to establish a sliding window, employing triangulation to calculate the initial depths of image feature points in window. And sliding window otpimization is adopted to update the depth of feature points until the update value less than a threshold. To guarantee full parallax between image feature points within the sliding window, a method based on the epipolar geometric constraints is introduced to recover feature points that may failed in optical flow tracking.
	\item A novel hybrid LiDAR-Visual data association method is introduced for localization. Initially, both projection points and feature points are utilized to establish the reprojection error. Subsequently, ESIKF is employed for accurate state estimation with the established reprojection error. By fully leveraging environmental data from both LiDAR and Visual to constrain state estimation, the proposed HDA-LVIO can achieve more accurate localization than adopting either the feature-based method or the direct method alone.
	\item Extensive experiments are conducted in urban environments utilizing data from KITTI datasets, NTU-VIRAL datasets and our platform. Simultaneously, the localization results are compared with numerous existing algorithms to validate the accuracy improvements achieved by the proposed HDA-LVIO algorithm.
\end{itemize}

\begin{figure*}[htbp]
	\centering
	\includegraphics[width=5.0in]{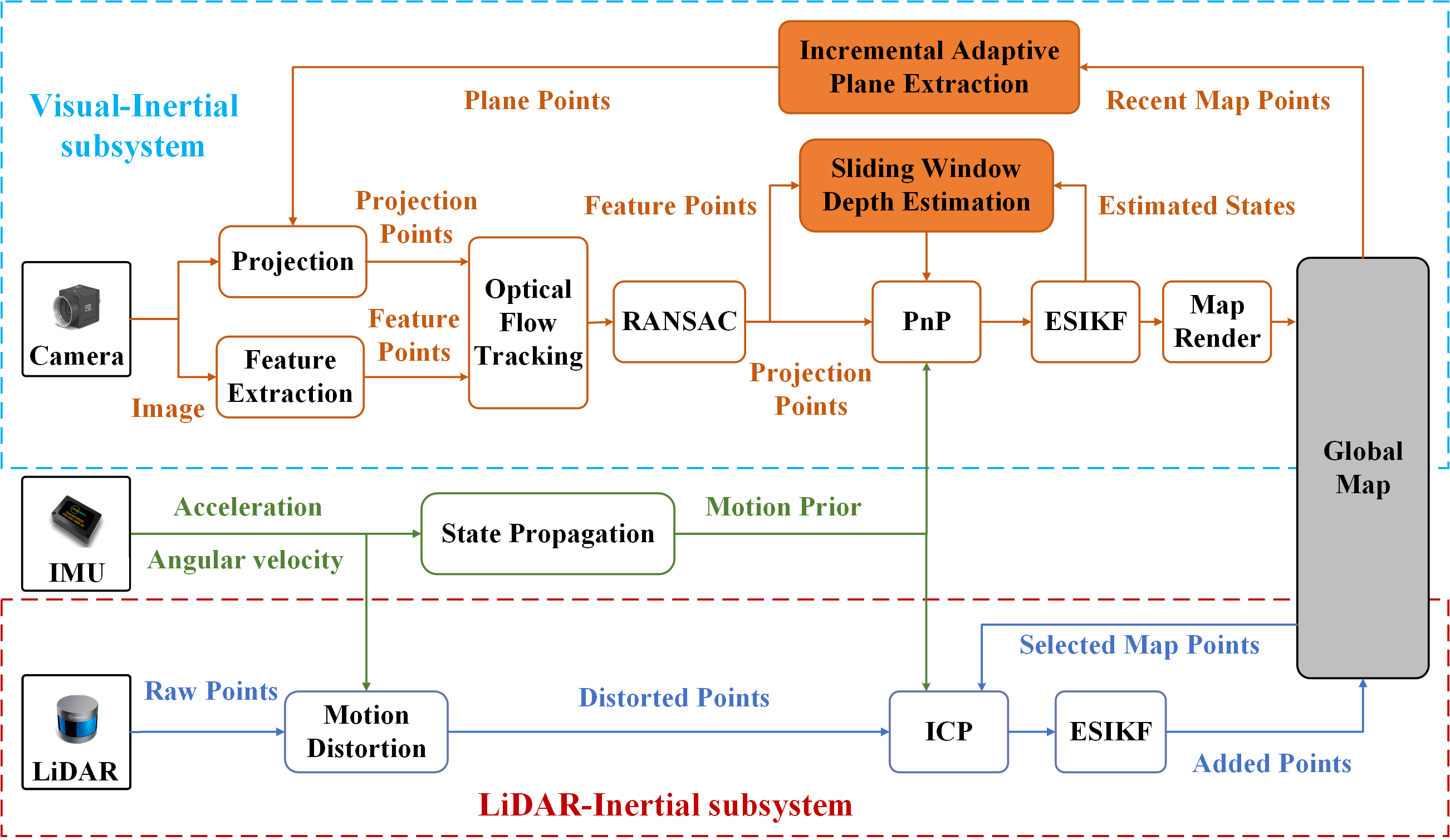}
	\caption{{The framework of the proposed HDA-LVIO.} }
	\label{framework}
\end{figure*}

\section{Related Notations}
The mathematics notations employed in this work are defined here. Bold letters represent vectors and matrices, while scalars are denoted by thin letters. The image frame is denoted by the symbol $b$, the world frame is represented by the letter $g$, and the LiDAR frame is represented by the letter $l$.  The pointcloud is represented with $\boldsymbol{P}$, and the 3D LiDAR point in pointcloud is represented with $\boldsymbol{p}$.
And $\left(\cdot\right)^{\wedge}$ represents the mapping from a vector to a skew symmetric matrix. The norm of vector $\boldsymbol{a}$  is represented as $\left\|\boldsymbol{a}\right\|$. The transpose of vector or matrix is desctribed with ${\left(\cdot\right)}^{\top}$.

The transformation matrix, denoted as $\boldsymbol{T} \in \mathbb{R}^{4\times4}$, encapsulates the relationship between coordinate systems. For example, $\boldsymbol{T}_a^b$ represents the transformation from frame $b$ to $a$. And $\boldsymbol{T}$ comprises both the rotation matrix $\boldsymbol{R}\in  \mathbb{R}^{3\times3}$ and the translation vector  $\boldsymbol{t}\in  \mathbb{R}^{3\times1}$. The specific form of transformation matrix $\boldsymbol{T}$ is as follows:
\begin{equation}
	\label{trans relation}
	\boldsymbol{T}=\begin{bmatrix}
		\boldsymbol{R}& \boldsymbol{t} \\
		0&1
	\end{bmatrix}
\end{equation}

And the matrix chain rule is as follows:
\begin{equation}
	\label{matrix chain rule}
	\boldsymbol{T}_a^c = \boldsymbol{T}_b^c \boldsymbol{T}_a^b
\end{equation}

For the vector $\boldsymbol{t}$, the homogeneous coordinate of $\boldsymbol{t}$ is written as follows:
\begin{equation}
	\label{homogeneous}
	\bar{\boldsymbol{t}} = [\boldsymbol{t}, 1] \in \mathbb{R}^{4\times1}
\end{equation}

\section{LiDAR Visual Inertial Odometry with Hybrid Data Association }
\subsection{Overview}
The framework of the proposed HDA-LVIO is illustrated in Fig. \ref{framework}. The system consists of two parts, namely the Visual-Inertial subsystem (VIS) and the LiDAR-Inertial subsystem (LIS). For the LIS, firstly, motion distortion in the LiDAR pointcloud is eliminated with IMU data. Subsequently, Iterative Closest Point (ICP) errors is constructed by registrating LiDAR scan to the global map. And then Error State Iterated  Kalman Filter (ESIKF) is employed for state estimation. For detailed information about the LIS, please refer to \cite{xu2022fast}. 


For the VIS, the recent global map is first voxelized, and an incremental adaptive method is adopted to extract planes from each voxel. The centroids of these planes are then projected onto the image to obtain projection points, and Harris feature points are extracted from the image. Associations between adjacent image frames are established by applying optical flow tracking to both the projection points and feature points. Subsequently, RANSAC is employed to eliminate erroneous tracking points. The depth of feature points are estimated using sliding window optimization. A method based on epipolar geometric constraints is proposed to recover feature points when optical flow tracking fails, ensuring significant parallax of feature points within the sliding window. Finally, utilizing accurately tracked feature points and projection points, the reprojection error is constructed, and state estimation is achieved through ESIKF.

\subsection{Incremental Adaptive Plane Extraction}
\label{Incremental Adaptive Plane Extraction}
\begin{figure}[htbp]
	\centering
	\includegraphics[width=2.5in]{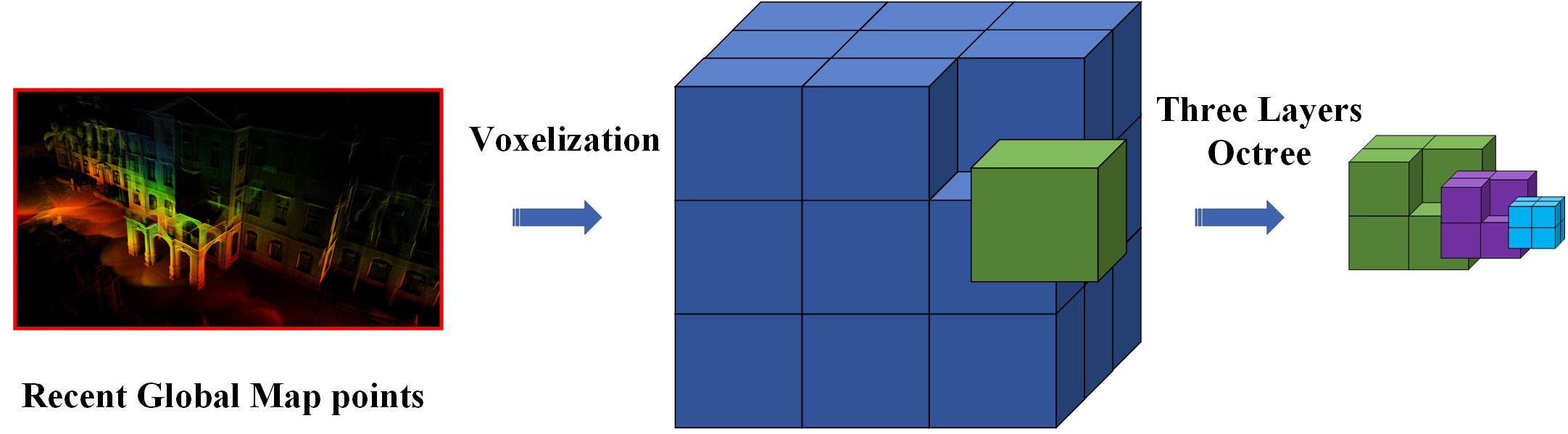}
	\caption{{The flowchart of the proposed plane extraction method.} }
	\label{plane select}
\end{figure}

In this section, the proposed method for extracting planes from the recent global map points $\boldsymbol{P}^{g}$ is explained in detail. Using the states estimated by the LIS, the $i$-th frame of LiDAR pointcloud $\boldsymbol{P}_{i}^l$ can be transformed to the $g$ frame and added to the recent global map $\boldsymbol{P}^{g}$. And $\boldsymbol{P}_{i}^g$ is divided into 3D voxels $\left\{V_q, q=1,...,Q\right\}$ with a size of 1 meter, and a hash table is employed to manage these voxels.
And then the three layers octree  is employed to adaptively extract plane from these voxels.
 The flowchart of the proposed method is depicted as Fig. \ref{plane select}.

Once the number of pointcloud $\boldsymbol{P}_{q}^g$ within the $q$-th voxel $V_q$ surpasses a predefined threshold $t_s$, the initialization of the pointcloud within that voxel will be operated. Firstly, the mean $\bar{\boldsymbol{p}}$ and covariance matrix $\rm{cov}_{\boldsymbol{p}}$ of the points within the voxel are calculated  as follows:
\begin{equation}
	\label{mean pt}
	\bar{\boldsymbol{p}}_q = \frac{1}{N}\sum_{k = 1}^{N}\boldsymbol{p}_k^g
\end{equation}
\begin{equation}
	\label{covariance pt}
	\rm{cov}_{q} = \frac{1}{N}\sum_{k=1}^{N}\left(\boldsymbol{p}_k^g - \bar{\boldsymbol{p}}\right)\left(\boldsymbol{p}_k^g - \bar{\boldsymbol{p}}\right)^{T}
\end{equation}
where $N$ is the size of points in the $q$-th voxel $V_q$, and $\boldsymbol{p}_k^g$ is the $k$-th point in the $\boldsymbol{P}_{q}^g$.
 
The eigenvalues $\{\lambda_{i}, i = 1,2,3\}$ of matrix $\rm{cov}_{q}$ can be obtained through eigen decomposition. Once the smallest eigenvalue $\lambda_{min}$ meet the follow planar criteria depicted in Eq. \eqref{plane threshold}, the pointcloud within the voxel is recognized as part of the same plane. 

\begin{equation}
	\label{plane threshold}
	\lambda_{min}=\min{\{\lambda_{i}, i = 1,2,3\}} < t_{\lambda}
\end{equation}
where $t_{\lambda}$ is the threshold and set as $0.01m$ in this work.

Conversely, for cases where it does not meet the planar criteria  depicted in Eq. \eqref{plane threshold}, a three-layer octree is applied to partition the  $\boldsymbol{P}_{q}^g$, and a cyclic evaluation of the pointcloud within the child nodes is conducted to determine their conformity to the planar criteria depicted in Eq. \eqref{plane threshold}. The terination criteria of the cyclic evaluation encompass three conditions: 
\begin{itemize}
\item The points within the child node are below the specified threshold $t_{s}$. 
\item The pointcloud within the child node satisfies the plane threshold condition depicted in Eq. \eqref{plane threshold}. 
\item The child node is at the top level of the octree.
\end{itemize}

After the initialization for the voxel $V_q$, subsequently added points $\check{\boldsymbol{P}}_{q}^g$ will be used to update the mean $\bar{\boldsymbol{p}}_q$  and covariance $\rm{cov}_{q}$ of plane in the voxel $V_q$  or nodes in the three-layer octree as follows:
\begin{equation}
	\label{update plane mean}
	\check{\boldsymbol{p}} = \frac{1}{N+M}\left(N \bar{\boldsymbol{p}} + \sum_{k=1}^{M}\boldsymbol{p}_k^g\right)
\end{equation}

\begin{equation}
	\label{update palne cov}
	\check{\rm{cov}}_{q} = \frac{1}{N+M}\left(N \rm{cov}_{q} + \sum_{k=1}^{M}\left(\boldsymbol{p}_k^g - \check{\boldsymbol{p}}\right)\left(\boldsymbol{p}_k^g - \check{\boldsymbol{p}}\right)^{T}\right)
\end{equation}
where $N$ is the size of $\check{\boldsymbol{P}}_{q}^g$, $M$ is the size of new add points. 

It is essential to highlight that the update process described in Eq. \eqref{update plane mean} and \eqref{update palne cov} are only executed when the size of newly added pointcloud surpasses the specified threshold. 
\begin{algorithm}[H]
	\label{algorithm1}
	\caption{Incremental adaptive plane extraction algorithm} 
	\hspace*{0.02in} {\bf Input:} 
	The $i$-th frame of LiDAR pointcloud, $\boldsymbol{P}_{i}^l$,\\
	\hspace*{0.49in}The 3D voxels in Hash table, $\left\{V_q, q=1,...,Q\right\}$.\\
	\hspace*{0.02in} {\bf Output:} 
	The center point $\left\{\boldsymbol{p}_{c_t}, t=1,...,T\right\}$ of planes in \\
	\hspace*{0.56in} recent global map, $\boldsymbol{P}^{g}$. 
	\begin{algorithmic}[1]
		\State With the states output from the LIS, the $i$-th frame LiDAR pointcloud $\boldsymbol{P}^{l}_i$ can be transformed to the $\boldsymbol{P}^{g}_i$.
		\State {Voxelize the $\boldsymbol{P}_{i}^g$ and divide it into 3D voxels $\left\{V_s,  s=1,...,S\right\}$.}
		\For{each voxel $V_s$ in $\left\{V_s,  s=1,...,S\right\}$}
		\If{$V_s$ not exist in Hash table $\left\{V_q\right\}$}
		\State{Add $V_s$ into Hash table $\left\{V_q\right\}$.}
		\EndIf
		\If{the pointcloud $\boldsymbol{P}_{q}^g$ within $V_q$ is initialized}
		\State Update the plane as Eq. \eqref{update plane mean} and \eqref{update palne cov}.
		\State Calculate $\boldsymbol{p}^{\Delta}_q$ as Eq. \eqref{update delta mean}.
		\If{$\boldsymbol{p}^{\Delta}_q < t_{\Delta}, \check{\lambda}_{min} < t_{\lambda}$}			
		\State {Add the update mean  $\check{\boldsymbol{p}}$ into  $\left\{\boldsymbol{p}_{c_t}\right\}$.}
		\ElsIf{$\boldsymbol{p}^{\Delta}_q < t_{\Delta}, \check{\lambda}_{min} > t_{\lambda}$}
		\State {Add the initial mean  $\bar{\boldsymbol{p}}$ into  $\left\{\boldsymbol{p}_{c_t}\right\}$.}
		\ElsIf{$\boldsymbol{p}^{\Delta}_q > t_{\Delta}$}
		\State {Remove $V_q$ from $\left\{V_q\right\}$}
		\EndIf
		\ElsIf{the size of $\boldsymbol{P}_{q}^g > t_s$}
		\State Calculate the mean $\bar{\boldsymbol{p}}$ and covariance matrix 
		\hspace*{0.2in} $\rm{cov_{q}}$ 	as  Eq.  \eqref{mean pt} and \eqref{covariance pt}, and calculate  eigenvalues 
		\hspace*{0.2in} $\left\{\lambda_i, i=1,2,3 \right\}$ of $\rm{cov_{q}}$.
		\If {minimal eigenvalue $\lambda_{min} < t_{\lambda}$ }
		\State {Add the mean $\bar{\boldsymbol{p}}$ of pointcloud $\boldsymbol{P}_{q}^g$ into $\left\{\boldsymbol{p}_{c_t}\right\}$}.		
		\Else
		\State	Apply three-layer octree to partition the point 
		\hspace*{0.40in} cloud in this voxel and to obtain the pointcloud
		\hspace*{0.40in} in each child node.
		\State	Conduct the cylic planar evaluation of the 
		\hspace*{0.4in} pointcloud in each child nodes of the octree.	
		\If{(the  pointcloud within the child node fullfil   
			\hspace*{0.7in} the criteria depticted in Eq. \eqref{plane threshold})}
		\State Add the mean $\bar{\boldsymbol{p}}$ into  $\left\{\boldsymbol{p}_{c_t}\right\}$, and exit the 
		\hspace*{0.7in} cylic planar evaluation.
		\EndIf
		\EndIf
		\EndIf
		\EndFor
	\end{algorithmic}
\end{algorithm}
And the mean update value $\boldsymbol{p}^{\Delta}_q$ can be obtained as follows
\begin{equation}
	\label{update delta mean}
	\boldsymbol{p}^{\Delta}_q = \big\|\check{\boldsymbol{p}} - \bar{\boldsymbol{p}}  \big\|
\end{equation}

Simultaneously, the eigenvalues $\{\check{\lambda}_{i}, i = 1,2,3\}$ of updated covariance matrix $\check{\rm{cov}}_{q}$ can be obtained through eigen decomposition.

Subsequently, the plane updation criteria is defined as follows:
\begin{equation}
	\label{update criteria}
	\boldsymbol{p}^{\Delta}_q < t_{\Delta}, \check{\lambda}_{min} < t_{\lambda}
\end{equation}
where $t_{\Delta}$ and $t_{\lambda}$ are the thresholds and separtely set as $0.1m$ and $0.01m$.

If both $\boldsymbol{p}_q^{\Delta}$ and $\check{\lambda}_{min}$ fullfil the criteria written in Eq. \eqref{update criteria}, the mean and covariance of the plane in voxel $V_q$ will be updated as $\check{\boldsymbol{p}}$ and $\check{\rm{cov}}_{q}$.
Otherwise, it indicates that the plane within the voxel has changed, which corresponds to the following two situations. 
\begin{itemize}
	\item $\boldsymbol{p}^{\Delta}_q > t_{\Delta}$. Dynamic pointcloud exist in this voxel, which may be caused by moving vehicles. The voxel $V_q$ is removed from $\left\{V_q\right\}$.
	\item $\boldsymbol{p}^{\Delta}_q < t_{\Delta}, \check{\lambda}_{min} > t_{\lambda}$. The accuracy of global map constructed by the LIS system decreases. And the mean and covariance of the plane in voxel $V_q$ is still set as  $\bar{\boldsymbol{p}}$ and ${\rm{cov}}_{q}$.
\end{itemize}

The detail procession of the proposed incremental adaptive plane extraction algorithm is depicted in Algorithm 1.

\subsection{Feature Point Depth Estimation}
\label{Feature Point Depth Recovery}
\begin{figure}[htbp]
	\centering
	\includegraphics[width=3.4in]{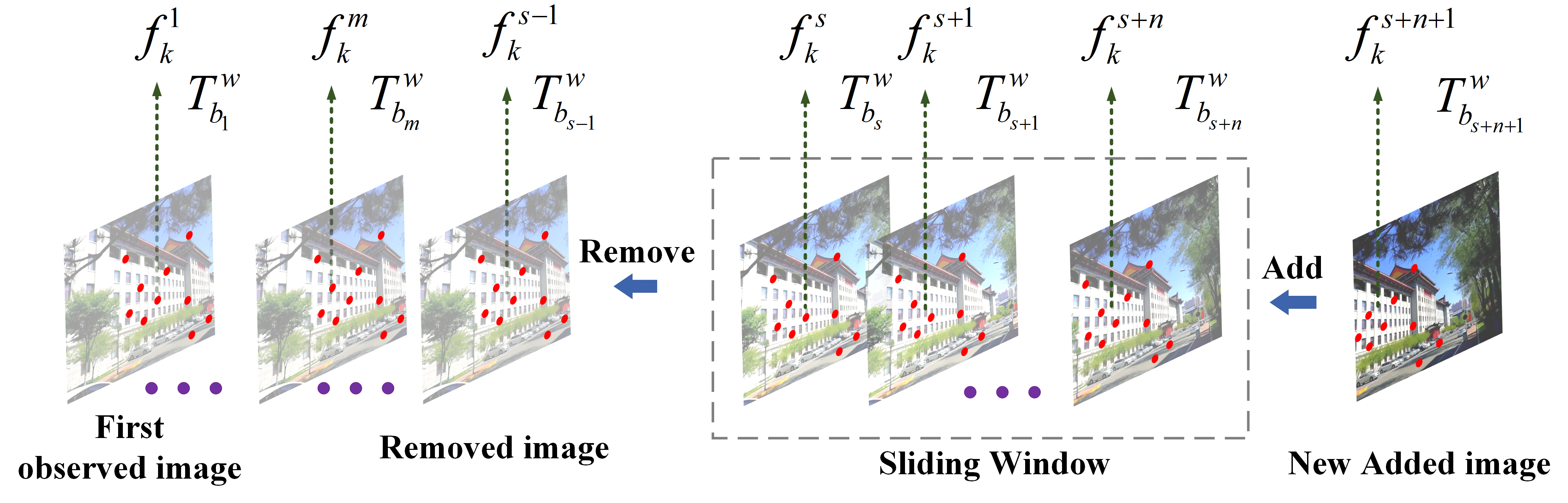}
	\caption{{Image frame utilized in feature point depth recovery.} }
	\label{sliding window}
\end{figure}

In the proposed feature point depth estimation algorithm, we initially obtain the depth of the feature point through triangulation. Subsequently, the depth of feature point is updated via sliding window optimization. To guarantee full parallax between image feature points within the sliding window, we introduce the method based on epipolar geometric constraints to recover feature points that may failed in optical flow tracking. The image frame utilized for feature point depth recovery is depicted in Fig. \ref{sliding window}, where the red dots are the feature points in the images. We take the $k$-th feature point $\boldsymbol{f}_k$ as an example to illustrate the proposed depth estimation algorithm.

\subsubsection{feature point recovery}
\begin{figure}[htbp]
	\centering
	\includegraphics[width=2.5in]{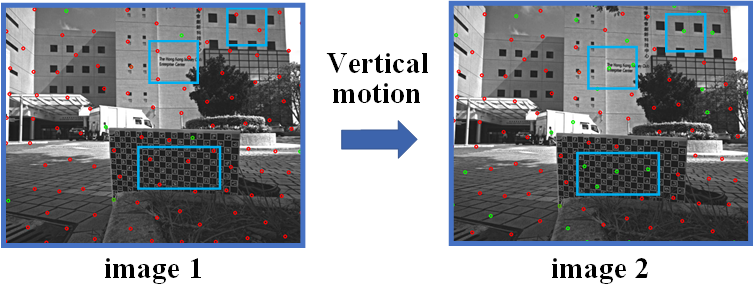}
	\caption{{Feature point tracking failed due to rapid vertical movement.} }
	\label{error track}
\end{figure}

The depth of feature point $\boldsymbol{f}_k$ is optimizated within the sliding window. To ensure the accuracy of optimization, obtaining an adequate parallax for feature point $\boldsymbol{f}_k$ within the sliding window is crucial. However, in cases of rapid motion or the presence of repetitive textures within the actual scene, feature points may not be consistently tracked over an extended period, which will pose challenges in obtaining substantial parallax for depth recovery. As illustrated in Fig. \ref{error track}, the accurately tracked feature points are represented by the red points, while the green points denote new added feature points to replace the error tracking feature points in image 2. The failure in tracking feature points within the blue box will be attributed to the rapid vertical movement. 

To recover the lost feature point $\boldsymbol{f}_k^n$ in the $n$-th image frame during tracking, the coarse feature point $\tilde{\boldsymbol{f}}{_k^n}$ is first calculated based on the epipolar geometric constraints. And then the $\tilde{\boldsymbol{f}}{_k^n}$ is optimizated with pyraid optical flow tracking to obtain the accurate $\boldsymbol{f}_k^n$. 
By employing the recent $n$ image poses $\left\{ {\boldsymbol{T}}^{w}_{b_i}, i = 1,...,n\right\}$ and the matrix chain rule expressed in Eq. \eqref{matrix chain rule}, ${\boldsymbol{T}}_{{b_n}}^{{b_{i}}}$ can be obtained as follows:
\begin{equation}
	\label{cicn}
{\boldsymbol{T}}_{{b_n}}^{{b_{i}}} =\left({\boldsymbol{T}}^{{w}}_{{b_{i}}}\right)^{\mathrm{T}}{\boldsymbol{T}}^{{w}}_{{b_{n}}} 
\end{equation}
where ${\boldsymbol{T}}_{{b_n}}^{{b_{i}}}  =\begin{bmatrix}
		\boldsymbol{R}_{{b_n}}^{{b_{i}}}& \boldsymbol{t}_{{b_n}}^{{b_{i}}} \\
		0&1 \end{bmatrix}$.

Along with feature points $\left\{\boldsymbol{f}_k^i, i=1,...,n-1\right\}$, we can obtain $\boldsymbol{f}_k^n$ through the epipolar geometric constraints as follows:
\begin{equation}
	\label{epipolar constraints}
	\left\{ \begin{split}
			&\left({{\boldsymbol{\rm{K}}}^{ - 1}}{\bar{\boldsymbol{f}}_k^1}\right)^{\mathrm{T}}   {{\left( {\boldsymbol{t}_{b_n}^{b_1}} \right)}^ \wedge }\boldsymbol{R}_{b_n}^{b_1}{{\boldsymbol{\rm{K}}}^{ - 1}}{\bar{\boldsymbol{f}}_k^n} = 0\\
			&\left({{\boldsymbol{\rm{K}}}^{ - 1}}{\bar{\boldsymbol{f}}_k^2}\right)^{\mathrm{T}}   {{\left( {\boldsymbol{t}_{b_n}^{b_2}} \right)}^ \wedge }\boldsymbol{R}_{b_n}^{b_2}{{\boldsymbol{\rm{K}}}^{ - 1}}{\bar{\boldsymbol{f}}_k^n} = 0\\
			&\hspace{25mm}\vdots  \\
			&\left({{\boldsymbol{\rm{K}}}^{ - 1}}{\bar{\boldsymbol{f}}_k^{n-1}}\right)^{\mathrm{T}}   {{\left( {\boldsymbol{t}_{b_n}^{b_{n-1}}} \right)}^ \wedge }\boldsymbol{R}_{b_n}^{b_{n-1}}{{\boldsymbol{\rm{K}}}^{ - 1}}{\bar{\boldsymbol{f}}_k^n} = 0
	\end{split} \right.
\end{equation}
where ${\boldsymbol{\rm{K}}}$ represents the intrinsic parameters of camera, ${\bar{\boldsymbol{f}}_k^n}$ is the homogeneous coordinate form of ${{\boldsymbol{f}}_k^n}$ as depicted in Eq. \eqref{homogeneous}.

The above equation can be simplified to the following form:
\begin{equation}
	\boldsymbol{A}{\bar{\boldsymbol{f}}_k^n} = \left[\boldsymbol{A}_1, \boldsymbol{A}_2, \boldsymbol{A}_3\right]\left[{{\boldsymbol{f}}_k^n}, 1\right]^{\top}  = \boldsymbol{0}
\end{equation}
where
\begin{equation}
	\boldsymbol{A}=\left[ {\begin{array}{*{20}{c}}
			\left({{\boldsymbol{\rm{K}}}^{ - 1}}{\bar{\boldsymbol{f}}_k^1}\right)^{\mathrm{T}}   {{\left( {\boldsymbol{t}_{b_n}^{b_1}} \right)}^ \wedge }\boldsymbol{R}_{b_n}^{b_1}{{\boldsymbol{\rm{K}}}^{ - 1}}\\
			\left({{\boldsymbol{\rm{K}}}^{ - 1}}{\bar{\boldsymbol{f}}_k^2}\right)^{\mathrm{T}}   {{\left( {\boldsymbol{t}_{b_n}^{b_2}} \right)}^ \wedge }\boldsymbol{R}_{b_n}^{b_2}{{\boldsymbol{\rm{K}}}^{ - 1}}\\
			\vdots \\
			\left({{\boldsymbol{\rm{K}}}^{ - 1}}{\bar{\boldsymbol{f}}_k^{n-1}}\right)^{\mathrm{T}}   {{\left( {\boldsymbol{t}_{b_n}^{b_{n-1}}} \right)}^ \wedge }\boldsymbol{R}_{b_n}^{b_{n-1}}{{\boldsymbol{\rm{K}}}^{ - 1}}
	\end{array}} \right]\in\mathbb{R}^{(n-1)\times3}
\end{equation}

And then by rearranging the formula mentioned above, we can obtain:
\begin{equation}
	\boldsymbol{X}{{\boldsymbol{f}}_k^n}= \boldsymbol{y}
\end{equation}
where $\boldsymbol{X} = \left[\boldsymbol{A}_1, \boldsymbol{A}_2\right]\in\mathbb{R}^{(n-1)\times2}$, $ \boldsymbol{y} = -\boldsymbol{A}_3\in \mathbb{R}^{(n-1)\times1}$, and $\boldsymbol{A}_1, \boldsymbol{A}_2, \boldsymbol{A}_3$  correspond to the first, second, and third column of matrix $\boldsymbol{A}$.

Hence, we can establish the following objective function to estimate ${{\boldsymbol{f}}_k^n}$.

\begin{equation}
	\label{least square}
	\min_{{{\boldsymbol{f}}_k^n}}	\frac{1}{2}\Big\|\left(\boldsymbol{X}{{\boldsymbol{f}}_k^n} - \boldsymbol{y}\right)\Big\|^2
\end{equation}

However, considering the estimation error in $\left\{{\boldsymbol{T}}^{{w}}_{{b_{i}}}\right\}$, the ${\boldsymbol{T}}_{{b_n}}^{{b_{i}}}$ calculated with Eq.  \eqref{cicn} will also contain error. The weighted least squares method is utilized to solve the optimal solution for Eq. \eqref{least square}, aiming to reduce the impact of this error on the estimation result of ${{\boldsymbol{f}}_k^n}$. To obtian the weight $\left\{{w}_{i}, i = 1,...,n-1\right\}$ utilized in the weighted least squares method,  we firstly reproject the feature points $\left\{\boldsymbol{f}_k^i, i=1,...,n-2\right\}$ onto the $(n-1)$-th image frame $b_{n-1}$ to obtain the reprojection error $\left\{{e}_{i}, i=1,...,n-2\right\}$ as follow. 
\begin{equation}
	\label{reprojection error for weight}
	{e}_{i} = \left\| \left({{\boldsymbol{\rm{K}}}^{ - 1}}{\bar{\boldsymbol{f}}_k^i}\right)^{\mathrm{T}}   {{\left( {\boldsymbol{t}_{b_{n-1}}^{b_i}} \right)}^ \wedge }\boldsymbol{R}_{b_{n-1}}^{b_i}{{\boldsymbol{\rm{K}}}^{ - 1}}{\bar{\boldsymbol{f}}_k^{n-1}}\right\|
\end{equation}

The reprojection error ${e}_{i}$ is then utilized to form weight $\left\{w_i\right\}$ as follows.
\begin{algorithm}[H]
	\label{algorithm2}
	\caption{Feature Point Recovery Algorithm} 
	\hspace*{0.02in} {\bf Input:} 
	The recent image poses, $\left\{{\boldsymbol{T}}^{{w}}_{{b_{i}}}, i = 1,...,n\right\}$, \\
	\hspace*{0.3in}Feature point $\boldsymbol{f}_k$ in recent image,
	$\left\{\boldsymbol{f}_k^i, i=1,...,n-1\right\}$.\\
	\hspace*{0.02in} {\bf Output:} 
	The feature point $\boldsymbol{f}_k$ in the $n$-th image, $\boldsymbol{f}_k^n$.
	\begin{algorithmic}[1]
		\State Calculate the $\left\{{\boldsymbol{T}}_{{b_n}}^{{b_{i}}}, i = 1,...,n-1\right\}$ with  image poses $\left\{{\boldsymbol{T}}_{{w}}^{{b_{i}}}, i = 1,...,n\right\}$ as Eq. \eqref{cicn}.
		\State Establish the epipolar geometric constraints as Eq. \eqref{epipolar constraints} with $\left\{{\boldsymbol{T}}_{{b_n}}^{{b_{i}}}, i = 1,...,n-1\right\}$ and the corresponding feature points $\left\{\boldsymbol{f}_k^i, i=1,...,n-1\right\}$.
		\State Calculate the reprojection error $\left\{{e}_{i}\right\}$ as Eq. \eqref{reprojection error for weight}.
		\State Calculate the weight $\left\{{w}_{i}\right\}$ as Eq. \eqref{weight}.
		\State Obtain the weighted least squares solution ${\tilde{\boldsymbol{f}}{_k^n}}$ as Eq. \eqref{least square solution}.
		\State Use ${\tilde{\boldsymbol{f}}{_k^n}}$ as the initial value for pyramid optical flow tracking to obtain ${{\boldsymbol{f}}_k^n}$.
	\end{algorithmic}
\end{algorithm}
\begin{equation}
	\label{weight}
	\left\{w_i = \frac{e_i}{{e_{\min}}} + 1, i = 1,...,n-2 \right\}
\end{equation}
where 
	$e_{\min} = \min \left\{e_i, i = 1,...,n-2\right\}$.
Specifically, since the image frame $b_{n-1}$ is more close to the image frame $b_n$, the error associated with the transformation matrix ${\boldsymbol{T}}_{{b_n}}^{{b_{n-1}}}$ is relatively smaller compared to transformation matrices  $\left\{{\boldsymbol{T}}_{{b_n}}^{{b_{i}}}, i=1,...,n-2\right\}$. Therefore, $w_{n-1}$ is set as 1. And the weight matrix $\boldsymbol{W}$ can be obtained with $\left\{w_i, i=1,...,n-1\right\}$ as follows.

\begin{equation}
	\boldsymbol{W}=\begin{bmatrix}
		w_1&  0& 0\\
		0&  \ddots &0 \\
		0&  0&w_{n-1}
	\end{bmatrix}
\end{equation}

Therefore, we can obtain the weighted least squares solution for ${{\boldsymbol{f}}_k^n}$ as follows.
\begin{equation}
	\label{least square solution}
	{\tilde{\boldsymbol{f}}{_k^n}} = \left(\boldsymbol{X}^T\boldsymbol{W}\boldsymbol{X}\right)^{-1}\boldsymbol{X}^{T}\boldsymbol{W}\boldsymbol{y}
\end{equation}


\begin{figure}[htbp]
	\centering
	\includegraphics[width=2.2in]{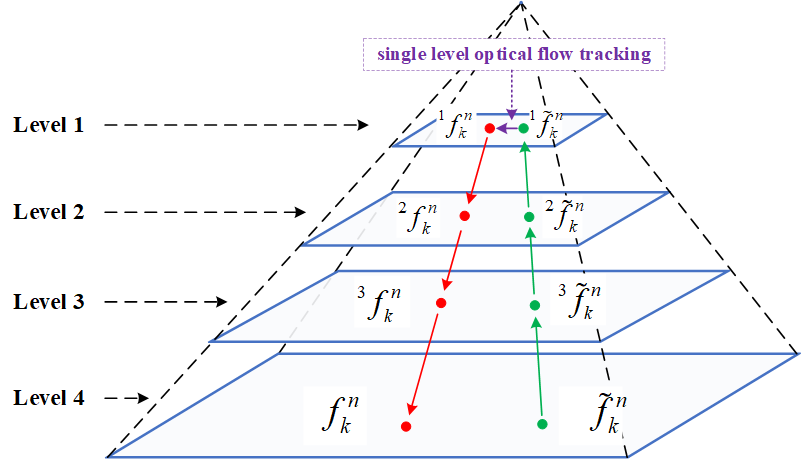}
	\caption{{Feature point depth recovery with pyraid optical flow tracking.} }
	\label{feature tracking}
\end{figure}

And then, we use ${\tilde{\boldsymbol{f}}{_k^n}}$ as the initial value for pyramid optical flow tracking to calculate the accurate $\boldsymbol{f}_k^n$. The process of pyramid optical flow tracking is illustrated in Fig. \ref{feature tracking}. Firstly, we successively scale the $n$-th image frame $b_m$ three times with a scaling ratio of 0.5 to generate a four-level image pyramid \cite{adelson1984pyramid}. Then, the feature point ${\tilde{\boldsymbol{f}}{_k^n}}$ in the level 1 of the image pyramid $^1{\tilde{\boldsymbol{f}}{_k^n}}$ is obtained and the single level optical flow tracking proposed in \cite{lucas1981iterative} is adopted to calculate the ${^1}{{\boldsymbol{f}}_k^n}$.
And the accurate ${{\boldsymbol{f}}_k^n}$ can be obtained as follows.
\begin{equation}
	{{\boldsymbol{f}}_k^n} = 2^3 \cdot {^1{{\boldsymbol{f}}_k^n}}
\end{equation}

The detail process of the feature point recovery algorithm is listed in Algorithm 2.

\subsubsection{triangulation}
\label{triangulation}
In the context of triangulation, we use the image frame $b_1$ in which the feature point $\boldsymbol{f}_k$ is firstly detected as the reference, and subsequently calculate the parallax ${X}_{1,m}$ between the image frame $b_1$  and the  subsequent image frame $b_m$ as follows,
\begin{equation}
	{X}_{1,m} = \Big\| \boldsymbol{f}_k^1 -  \boldsymbol{f}_k^m \Big\|
\end{equation}
where $\boldsymbol{f}_k^1$ is the $k$-th feature point firstly observed in  image frame $b_1$, and $\boldsymbol{f}_k^m$ is the $k$-th feature point observed in  image frame $b_m$.

When the parallax ${X}_{1,m}$ fulfill the criteria depicted as follows:

\begin{equation}
	\label{parallax criteria}
	{X}_{1,m} > t_X
\end{equation}
where $t_X$ is the threshold, and is set as $10$ pixels.
The triangulation proposed in \cite{qin2018vins} is implemented and the initial depth $\tilde{d}_{{f}{_k^m}}$ of $\boldsymbol{f}_k^m$ can be  calculated.
\subsubsection{feature point depth estimation}
We utilize sliding window optimization to obtain the accurate feature point depth ${d}_{{f}_k}$ based on the initial depth $\tilde{d}_{{f}_k^m}$ calculated with triangulation. Within the sliding window depicted in Fig.\ref{sliding window}, the first image frame $b_s$ in the sliding window is set as the reference frame. And the depth of $\boldsymbol{f}^s_k$, denoted as ${d}_{{f}_k^s}$, is estimated with sliding window optimization. The initial value of ${d}_{{f}_k^s}$, described as $\tilde{d}_{{f}_k^s}$, is calculated with $\tilde{d}_{{f}_k^m}$ as follows.
\begin{equation}
	\label{depth transform}
	\tilde{d}_{{f}_k^s}=\left(\boldsymbol{T}^w_{b_{s}}\right)^{\mathrm{T}}\boldsymbol{T}_{b_m}^w\tilde{d}_{{f}_k^m}\boldsymbol{\rm{K}}^{-1}{\boldsymbol{f}}{^m_k }
\end{equation}

Then the  reporjection error $\boldsymbol{r}\big({d}_{{f}_k^s}, i\big)$ can be represented as follows:
\begin{equation}
	\label{visual residual}
	\boldsymbol{r}\big({d}_{{f}_k^s}, i\big) = \boldsymbol{\rm{K}}\left(\boldsymbol{T}_{b_{i}}^w\right)^{\mathrm{T}}\boldsymbol{T}_{b_s}^w{d}_{{f}_k^s}\boldsymbol{\rm{K}}^{-1}{\boldsymbol{f}}{^s_k} - { \boldsymbol{f}}^i_k
\end{equation}
We employ the L-M optimization method \cite{more2006levenberg} to minimize the norm of $\boldsymbol{r}\big({d}_{{f}_k^s}, i\big)$ in order to obtain the Maximum Posteriori Estimation (MAP) result of ${d}_{{f}_k^s}$ as follows.
\begin{equation}
	\label{visual residual all}
	{\hat{d}_{{f}_k^s}}=\min_{{d}_{{f}_k^s}}  \sum_{i=s+1}^{s+n}\left\| \boldsymbol{r}\big({d}_{{f}_k^s}, i\big) \right\|^2
\end{equation}
where ${\hat{d}_{{f}_k^s}}$ is the optimization result of ${d}_{{f}_k^s}$, and $n$ is the size of the image frame in sliding window.

\begin{algorithm}[H]
	\label{algorithm3}
	\caption{Feature Point Depth Estimation Algorithm} 
	\hspace*{0.02in} {\bf Input:} 
	The recent image states, $\left\{{\boldsymbol{T}}^{{w}}_{{b_{i}}}, i = 1,...,n\right\}$, \\
	\hspace*{0.3in}Feature point $\boldsymbol{f}_k$ in recent images,
	$\left\{\boldsymbol{f}_k^i, i=1,...,n\right\}$.\\
	\hspace*{0.02in} {\bf Output:} 
	The map point obtained with feature point $\boldsymbol{f}_k$, $\boldsymbol{p}_{f_k}$.
	\begin{algorithmic}[1]
		\State Adopt triangulation to calculate the initial depth $\tilde{d}_{{f}_k^m}$ as Sec. \ref{triangulation}.
		\State Calculate the $\tilde{d}_{{f}_k^s}$ with $\tilde{d}_{{f}_k^m}$ as Eq. \eqref{depth transform}.
		\Repeat
		\State Calculate the parallax $\textbf{X}_{s+n,s+n+1}$ of $\boldsymbol{f}_k$ between the 
		\hspace*{0.08in} newly added image frame $b_{s+n+1}$ and the latest frame 
		\hspace*{0.08in} $b_{s+n}$ in the sliding window.
		\If{$\textbf{X}_{s+n,s+n+1}$ fulfill the criteria depicted in Eq.
			\hspace*{0.25in} \eqref{par criteria}}
		\State Calculate the reporjection error $\boldsymbol{r}\big({d}_{{f}_k^s}, i\big)$ as 
		\hspace*{0.25in} Eq. \eqref{visual residual}.
		\State Obtain the MAP result of ${d}_{{f}_k^s}$ as 	Eq. \eqref{visual residual all}.
		\State Transform the estimated feature point depth in the 
		\hspace*{0.25in} sliding window as Eq. \eqref{feature point depht transform}.
		\State Remove the first image frame $b_s$ from the sliding 
		\hspace*{0.25in} window.
		\State Add the new image frame $b_{s+n+1}$ into the sliding 
		\hspace*{0.25in} window.
		\EndIf
		\Until{The update value of the feature point $\boldsymbol{f}_k$ in observed in the subsequent $u$-th image frame $b_u$ fulfill the criteria depicted in Eq. \eqref{update visual}}
		\State Obtain the map point $\boldsymbol{p}_{f_k}$ with the feautre point $\boldsymbol{f}_k^u$ as Eq. \eqref{map feature}.
	\end{algorithmic}
\end{algorithm}

When the parallax ${X}_{s+n,s+n+1}$ of $\boldsymbol{f}_k$ between the newly added image frame $b_{s+n+1}$ and the latest frame $b_{s+n}$ in the sliding window satisfies the following conditions,
\begin{equation}
	\label{par criteria}
	{X}_{s+n,s+n+1} > t_X
\end{equation}
the newly added image frame $b_{s+n+1}$ will be added into the sliding window. Subsequently, the first image frame $b_s$ are removed from the sliding window, and the estimated feature point depth in the sliding window is transformed as follow:
\begin{equation}
	\label{feature point depht transform}
	\tilde{d}_{{f}_k^{s+1}}=\left(\left(\boldsymbol{T}^w_{b_{s+1}}\right)^{\mathrm{T}}\boldsymbol{T}_{b_s}^w\hat{d}_{{f}_k^s}\boldsymbol{\rm{K}}^{-1}{\boldsymbol{f}}{^s_k }\right)_{\scaleto{z}{3.5pt}}
\end{equation}
where $\left(\cdot\right)_{z}$ depict the z-axis value of vector.

Similarly, the $\hat{d}_{{f}_k^{s+1}}$ can be obtained with Eq. \eqref{visual residual all}. 
As the image frame within the sliding window is updated, the depth of feature point $\boldsymbol{f}_k^s$ is continuously estimated using Eq. \eqref{visual residual all} and \eqref{feature point depht transform}.  If the depth update value $\delta{d}_{{f}_k^u}$ of feature point  in the subsequent $u$-th image frame fulfill the criteria depicted in Eq. \eqref{update visual}, the feature point depth estimation algorithm is considered complete.
\begin{equation}
	\label{update visual}
	\delta{d}_{{f}_k^u} = \left\|{\hat{d}_{{f}_k^u}} - \tilde{d}_{{f}_k^u} \right\| < t_d
\end{equation}
where $t_d$ is set as $0.1m$, ${\hat{d}_{{f}_k}}$ is the MAP result of ${d}_{{f}_k}$ obtained with Eq. \eqref{visual residual all}, and $\tilde{d}_{{f}_k}$ is the feature point depth obtained with Eq. \eqref{feature point depht transform}.

And the map point $\boldsymbol{p}_{f_k}$ can be obtained with the feautre point $\boldsymbol{f}_k^u$ as follows.
\begin{equation}
	\label{map feature}
	\boldsymbol{p}_{f_k}=	\boldsymbol{T}_{b_u}^w\hat{d}_{{f}_k^u}\boldsymbol{\rm{K}}^{-1}{\boldsymbol{f}}{^u_k }
\end{equation}

And the detail of  proposed feature depth estimation algorithm is listed in Alorithm 3.

\subsection{ESIKF}
\label{ESIKF}
We employ the ESIKF \cite{he2021kalman} to estimate the state $\boldsymbol{x}_i$ in VIS. And $\boldsymbol{x}_i$  is consisted with following elements:
\begin{equation}
	\boldsymbol{x}_i = \left[\boldsymbol{T}_{b_i}^w, \boldsymbol{v}_{b_i}^w, \boldsymbol{b}_a, \boldsymbol{b}_g\right]
\end{equation}
where $\boldsymbol{T}_{b_i}^w$ is the pose of image frame $b_i$, $\boldsymbol{v}_{b_i}^w$ is the velocity of the image frame $b_i$, $\boldsymbol{b}_a$ is the bias of the accelerometer in IMU, and $\boldsymbol{b}_g$ is the bias of the  gyroscope in IMU.

We simultaneously utilize map points $\left\{\boldsymbol{p}_{{f}_k}, k=1,...,K\right\}$ obtained in Sec. \ref{Feature Point Depth Recovery}  and plane center points $\left\{\boldsymbol{p}_{c_t}, t=1,...,T\right\}$ obtained in Sec. \ref{Incremental Adaptive Plane Extraction} for state estimation. First, the projection points $\left\{\boldsymbol{c}_t, t=1,...,T\right\}$ can be obtained from the plane center points $\left\{\boldsymbol{p}_{c_t}, t=1,...,T\right\}$ as follows:
\begin{equation}
	\boldsymbol{c}_t^{i-1} = \boldsymbol{\rm{K}}\left(\boldsymbol{T}_{b_{i-1}}^w\right)^{\mathrm{T}}\boldsymbol{p}_{c_t}, t=1,...,T
\end{equation}

And then, the feature points $\left\{\boldsymbol{f}_k^{i-1}, k=1,...,K\right\}$ 
and the projection points $\left\{\boldsymbol{c}_t^{i-1}, t=1,...,T\right\}$ obtained from plane center points are tracked with L-K optical flow to obtain the feature points $\left\{\boldsymbol{f}_k^{i}\right\}$ and projection points $\left\{\boldsymbol{c}_t^{i}\right\}$ in image frame $b_{i}$. Due to the photometric gradient of  the projection points $\left\{\boldsymbol{c}_t^{i-1}\right\}$ may not be readily apparent, the optical flow tracking results of these points  may exhibit errors. On the contrary, given the relatively clear photometric gradient of feature points $\left\{\boldsymbol{f}_k^{i-1}\right\}$, the optical flow tracking result of these points is more accurate and stable. To enhance the accuracy of optical flow tracking for projection points, we employ RANSAC \cite{chum2003locally} to simultaneously eliminate outliers in the tracking results of both feature points $\left\{\boldsymbol{f}_k^{i}\right\}$ and projection points $\left\{\boldsymbol{c}_t^{i}\right\}$. By leveraging the stable tracking of feature points to increase the proportion of effective tracking results in  both projection points and feature points, we improve the effectiveness of RANSAC in eliminating outliers from the tracking results of projection points $\left\{\boldsymbol{c}_t^{i}\right\}$. 

Subsequently, we construct the PnP reprojection error to update state $\boldsymbol{x}_i$. For the $k$-th feature point $\boldsymbol{p}_{{f}_k}$ and the $t$-th projection point $\boldsymbol{p}_{c_t}$, the reprojection error can be constructed as follows:
\begin{equation}
	\label{visual res1}
	\boldsymbol{0} = \boldsymbol{r}\left(\boldsymbol{x}_i, \boldsymbol{p}_{{f}_k}, \boldsymbol{f}_k^{i}\right) =    \textbf{K}\frac{1}{\boldsymbol{d}_{f_k^i}}\boldsymbol{T}_w^{b_i} \boldsymbol{p}_{{f}_k} - \boldsymbol{f}_k^{i}
\end{equation}
\begin{equation}
	\label{visual res2}
	\boldsymbol{0} = \boldsymbol{r}\left(\boldsymbol{x}_i, \boldsymbol{p}_{{c}_t}, \boldsymbol{c}_t^{i}\right) =    \textbf{K}\frac{1}{\boldsymbol{d}_{c_t^i}}\boldsymbol{T}_w^{b_i} \boldsymbol{p}_{{c}_t} - \boldsymbol{c}_t^{i}
\end{equation}
where $\boldsymbol{d}_{f_k^i}$ is the $z$-axis value of  $\boldsymbol{T}_w^{b_i} \boldsymbol{p}_{{f}_k}$, and $\boldsymbol{d}_{c_t^i}$ is the $z$-axis value of  $\boldsymbol{T}_w^{b_i} \boldsymbol{p}_{{c}_t}$

Due to the noise in $\boldsymbol{p}_{{f}_k} $, $\boldsymbol{f}_k^{i}$, and $\boldsymbol{p}_{{c}_t}$, $\boldsymbol{c}_t^{i}$,
\begin{equation}
	\begin{split}
		\tilde{\boldsymbol{p}}{_{{f}_k}} &= \boldsymbol{p}_{{f}_k} + \boldsymbol{n}_{{p}_{{f}}}, \hspace{2mm} \tilde{\boldsymbol{f}}{_k^{i}} = \boldsymbol{f}_k^{i} + \boldsymbol{n}_{{f}} \\
		\tilde{\boldsymbol{p}}{_{{c}_t}} &= \boldsymbol{p}_{{c}_t} + \boldsymbol{n}_{p_c}, \hspace{3.8mm} \tilde{\boldsymbol{c}}{_t^{i}} = \boldsymbol{c}_t^{i} + \boldsymbol{n}_{c}
	\end{split}
\end{equation}
and the estimation error $\delta \check{\boldsymbol{x}}_i$ between the posterior estimation $\check{\boldsymbol{x}}_i$ and the  state ${\boldsymbol{x}}_i$,
\begin{equation}
	\boldsymbol{x}_i = \check{\boldsymbol{x}}_i \boxplus \delta \check{\boldsymbol{x}}_i
\end{equation}
Eq. \eqref{visual res1} and \eqref{visual res2} can be rewritten as follows:
\begin{equation}
	\boldsymbol{0} = \boldsymbol{r}\Big(\check{\boldsymbol{x}}_i \boxplus \delta \check{\boldsymbol{x}}_i, \tilde{\boldsymbol{p}}{_{{f}_k}} - \boldsymbol{n}_{{p}_{{f}}}, \tilde{\boldsymbol{f}}{_k^{i}} - \boldsymbol{n}_{{f}}\Big)
\end{equation}
\begin{equation}
	\boldsymbol{0} = \boldsymbol{r}\Big(\check{\boldsymbol{x}}_i \boxplus \delta \check{\boldsymbol{x}}_i, \tilde{\boldsymbol{p}}{_{{c}_t}} - \boldsymbol{n}_{p_c}, \tilde{\boldsymbol{c}}{_t^{i}} - \boldsymbol{n}_{c}\Big)
\end{equation}
where $\boxplus$ is the manifold operateor introduced in \cite{xu2022fast}.

Then, the MAP estimation of $\delta \check{\boldsymbol{x}}_i$ can be obtained  as follows:
\begin{equation}
	\begin{split}
		\min_{\delta \check{\boldsymbol{x}}_i}\Big( &\Big\|\check{\boldsymbol{x}}_i \boxplus \delta \check{\boldsymbol{x}}_i \boxminus \hat{\boldsymbol{x}}_i + \mathcal{H}_i {\delta \check{\boldsymbol{x}}_i} \Big\|^2_{\Sigma_{\delta \hat{\boldsymbol{x}}_i}} + \Big. \\ \Big.
		&\sum_{k = 1}^{K}\Big\|\boldsymbol{r}\left(\check{\boldsymbol{x}}_i, \tilde{\boldsymbol{p}}{_{{f}_k}}, \tilde{\boldsymbol{f}}{_k^{i}}\right)  + \boldsymbol{\rm{H}}_{f_k}^i \delta \check{\boldsymbol{x}}_i \Big\|^2_{\Sigma_{\alpha_f}} + \Big. \\ \Big.
		&\sum_{t = 1}^{T}\Big\|\boldsymbol{r}\left(\check{\boldsymbol{x}}_i, \tilde{\boldsymbol{p}}{_{{c}_t}}, \tilde{\boldsymbol{c}}{_t^{i}} \right)  + \boldsymbol{\rm{H}}_{c_t}^i \delta \check{\boldsymbol{x}}_i \Big\|^2_{\Sigma_{\alpha_t}}\Big)
	\end{split}
\end{equation}
where $\hat{\boldsymbol{x}}_i$ in the prior state estimation of ${\boldsymbol{x}}_i$ obtained by state propagation \cite{lin2021r} with IMU data, $\Sigma_{\delta \hat{\boldsymbol{x}}_i}$ is the noise covariance of prediction error, $\Sigma_{\alpha_f}$ is the noise covariance of reprojection error calculated with feature points $\left\{\boldsymbol{f}_k^{i}, k=1,...,K\right\}$, $\Sigma_{\alpha_t}$ is the noise covariance of reprojection error calculated with projection points $\left\{\boldsymbol{c}_t^{i}, t=1,...,T\right\}$, and
\begin{equation}
	\mathcal{H}_i = {\bigg. {\frac{\partial\left(\check{\boldsymbol{x}}_i \boxplus \delta \check{\boldsymbol{x}}_i \boxminus \hat{\boldsymbol{x}}_i\right)}{\partial \delta \check{\boldsymbol{x}}_i}}\bigg|}_{\scaleto{\delta \check{\boldsymbol{x}}_i= 0}{7pt}} 
\end{equation}
\begin{equation}
	\boldsymbol{\rm{H}}_{f_k}^i = {\bigg. {\frac{\partial\boldsymbol{r}\big(\check{\boldsymbol{x}}_i\boxplus \delta \check{\boldsymbol{x}}_i, \tilde{\boldsymbol{p}}{_{{f}_k}}, \tilde{\boldsymbol{f}}{_k^{i}}\big)}{\partial \delta \check{\boldsymbol{x}}_i}}\bigg|}_{\scaleto{\delta \check{\boldsymbol{x}}_i= 0}{7pt}} 
\end{equation}
\begin{equation}
	\boldsymbol{\rm{H}}_{c_t}^i = {\bigg. {\frac{\partial\boldsymbol{r}\left(\check{\boldsymbol{x}}_i \boxplus \delta \check{\boldsymbol{x}}_i, \tilde{\boldsymbol{p}}{_{{c}_t}}, \tilde{\boldsymbol{c}}{_t^{i}} \right)}{\partial \delta \check{\boldsymbol{x}}_i}}\bigg|}_{\scaleto{\delta \check{\boldsymbol{x}}_i= 0}{7pt}}  
\end{equation}

Following \cite{xu2021fast}, the estimation state $\check{\boldsymbol{x}}_i$ can be obtained as follows:
\begin{equation}
	\check{\boldsymbol{x}}_i = \check{\boldsymbol{x}}_i \boxplus \left(-\textbf{K}_i\check{\textbf{z}}_i - \left(\textbf{I}-\textbf{K}_i \textbf{H}_i\right)\mathcal{H}_i^{-1}\left(\check{\boldsymbol{x}}_i \boxminus \hat{\boldsymbol{x}}_i\right)\right)
\end{equation}
where 
\begin{equation}
	\textbf{H}_i = \left[\boldsymbol{\rm{H}}_{f_1}^i,...,\boldsymbol{\rm{H}}_{f_K}^i,\boldsymbol{\rm{H}}_{c_1}^i,...,\boldsymbol{\rm{H}}_{c_T}^i\right]
\end{equation}
\begin{equation}
	\begin{split}
		\check{\textbf{z}}_i = \Big[\boldsymbol{r}\Big(\check{\boldsymbol{x}}_i, \tilde{\boldsymbol{p}}{_{{f}_1}}, \tilde{\boldsymbol{f}}{_1^{i}}\Big),...,\boldsymbol{r}\Big(\check{\boldsymbol{x}}_i, \tilde{\boldsymbol{p}}{_{{f}_K}}, \tilde{\boldsymbol{f}}{_K^{i}}\Big), \Big. \\ \Big.
		\boldsymbol{r}\Big(\check{\boldsymbol{x}}_i, \tilde{\boldsymbol{p}}{_{{c}_1}}, \tilde{\boldsymbol{c}}{_1^{i}} \Big),...,\boldsymbol{r}\Big(\check{\boldsymbol{x}}_i, \tilde{\boldsymbol{p}}{_{{c}_T}}, \tilde{\boldsymbol{c}}{_T^{i}} \Big)\Big]
	\end{split}
\end{equation}
\begin{equation}
	\textbf{K}_i = \left(\textbf{H}_i^{\textrm{T}}\textbf{R}^{-1}\textbf{H}_i + \textbf{P}^{-1}_i\right)^{-1}\textbf{H}_i^{\textrm{T}}\textbf{R}^{-1}
\end{equation}
\begin{equation}
	\textbf{R} = \textrm{diag}\left({\Sigma_{\alpha_f}},...,{\Sigma_{\alpha_f}},{\Sigma_{\alpha_c}},...,{\Sigma_{\alpha_c}}\right)
\end{equation}
\begin{equation}
	\textbf{P}_i = \left(\mathcal{H}_i\right)^{-1}\Sigma_{\delta \check{\boldsymbol{x}}_i}\left(\mathcal{H}_i\right)^{-\mathrm{T}}
\end{equation}


\section{Experiments and Discussions}
To validate the performance of the proposed HDA-LVIO in urban environments, a series of experiments are conducted utilizing the KITTI datasets \cite{geiger2013vision}, NTU-VIRAL datasets \cite{nguyen2022ntu}  and data collected from our platform. The KITTI dataset comprises data collected from vehicles, while the NTU-VIRAL dataset consists of data collected from aerial drones. These datasets exhibit diverse motion states, providing a comprehensive evaluation of  the proposed HDA-LVIO system. 

We select a set of  benchmark algorithms, namely LOAM\cite{zhang2014loam},  VINS-MONO \cite{qin2018vins}, FAST-LVIO\cite{zheng2022fast}, R3LIVE\cite{lin2022r},  and FAST-LIO\cite{xu2022fast} to demonstrate the superior localization accuracy of the proposed HDA-LVIO algorithm. Among these benchmark algorithms, LOAM is a LiDAR odometry, VINS-MONO is a Visual-Inertial odometry,  FAST-LIO is a LiDAR-Inertial odometry, while both R3LIVE and FAST-LVIO are LiDAR-Visual-Inertial odometry systems. And all experiments are conducted on a desktop computer equipped with an Intel Core i7-9700 processor running at 3.0GHz. The proposed HDA-LVIO and the benchmark algorithms are all operated in the Robot Operating System (ROS) \cite{quigley2009ros}, an open-source communication framework.

Absolute Trajectory Error (ATE) metric, as described in \cite{grupp2017evo}, are utilized to evaluate the localization accuracy. And the ATE metric can be computed as follows:
\begin{equation}
	\mathrm{ATE} = \sqrt{\frac{1}{N}\sum_{i=1}^{N}\left\|\boldsymbol{t}_{est}^i - \boldsymbol{t}_{gt}^i \right\|}
\end{equation}
where $\boldsymbol{t}_{est}^i$ is the  position estimated with odometry algorithm, and $\boldsymbol{t}_{gt}^i$ is the groundtruth position.

\subsection{NTU-VIRAL datasets}

\label{ntu-viral}

\begin{table*}[htbp]
	\centering
	\caption{{ATE of the comparison algorithms and the proposed HDA-LVIO in NTU-VIRAL dataset.}}
	\begingroup
	\setlength{\tabcolsep}{6pt} 
	\renewcommand{\arraystretch}{1.3} 
	\begin{tabular}{c|c|c|c|c|c|c}
		\Xhline{1.2px}
		\multicolumn{1}{c|}{\textbf{ATE (m)}} & \multicolumn{1}{c|}{{VINS-MONO}} & \multicolumn{1}{c|}{LOAM} & \multicolumn{1}{c|}{FAST-LIO}& \multicolumn{1}{c|}{R3LIVE}& \multicolumn{1}{c|}{FAST-LIVO}& \multicolumn{1}{c}{HDA-LVIO} \\
		\Xhline{1.0px}
		eee$\_$01 & 1.64 & 0.22 & 0.13 & 0.41 & 0.16 & \textbf{0.12}  \\
		\hline
		eee$\_$02 & {0.72} & {0.24} & {0.15} & {0.72} & {0.37} & \textbf{0.13} \\
		\hline
		eee$\_$03 & 1.05 & 0.17 & 0.16 & 0.92 & 0.41 & \textbf{0.14}  \\
		\Xhline{1.0px}
		nya$\_$01 & 1.48 & 0.25 & 0.12 & 0.42 & 0.29 & \textbf{0.09}  \\
		\hline
		nya$\_$02 & 0.58 & 0.31 & 0.14 & 0.42 & 0.16 & \textbf{0.12} \\
		\hline
		nya$\_$03 & 1.33 & 0.23 & \textbf{0.15} & 0.36 & 0.45 & \textbf{0.15} \\
		\Xhline{1.0px}
		sbs$\_$01 & 3.78 & 0.26 & 0.14 & 0.49 & 0.52 & \textbf{0.12}  \\
		\hline
		sbs$\_$02 & 1.54 & 0.19 & 0.14 & 0.47 & 1.10 & \textbf{0.10} \\
		\hline
		sbs$\_$03 & 1.30 & 0.50 & 0.13 & 0.35 & 0.17 & \textbf{0.08}  \\
		\Xhline{1px}
	\end{tabular}%
	\endgroup
	\label{ate for viral}%
\end{table*}%

\begin{table}[htbp]
	\centering
	\caption{The parameter of sensors utilized in NTU-VIRAL datasets.}
	\renewcommand{\arraystretch}{1.3} 
	\begin{tabular}{>{\setlength{\parskip}{1ex}}cc}
		\toprule
		\textbf{Sensor} & \textbf{Description} \\
		\midrule
		\makecell[c]{uEye \\ 1221 LE5} & \makecell[c]{a monochrome global-shutter camera,\\ with a resolution of 752\texttimes 480. }\\
		\addlinespace[0.5ex]
		\hline
		\addlinespace[0.5ex]
		\makecell[c]{VectorNav \\ VN1003} &\makecell[c]{ a MEMS IMU, with raw data rate of\\ 400Hz and gyro in-run bias $(5-7)^ {\circ} /\textrm{hr}$.}\\
		\addlinespace[0.5ex]
		\hline
		\addlinespace[0.5ex]
		\makecell[c]{Ouster OS1}  & \makecell[c]{a 16-channel ($-16^{\circ}, 16^{\circ}$)\\  LiDAR, rotating  at 10 Hz. }  \\
		\addlinespace[0.5ex]
		\hline
		\addlinespace[0.5ex]
		\makecell[c]{Leica \\ Nova MS60} & \makecell[c]{a scanning total station, \\providing  the  ground truth of  localization.} \\
		\bottomrule
	\end{tabular}%
	\label{NTU VIRAL SENSOR}%
\end{table}%

\begin{figure}[htbp]
	\centering
	\includegraphics[width=3.0in]{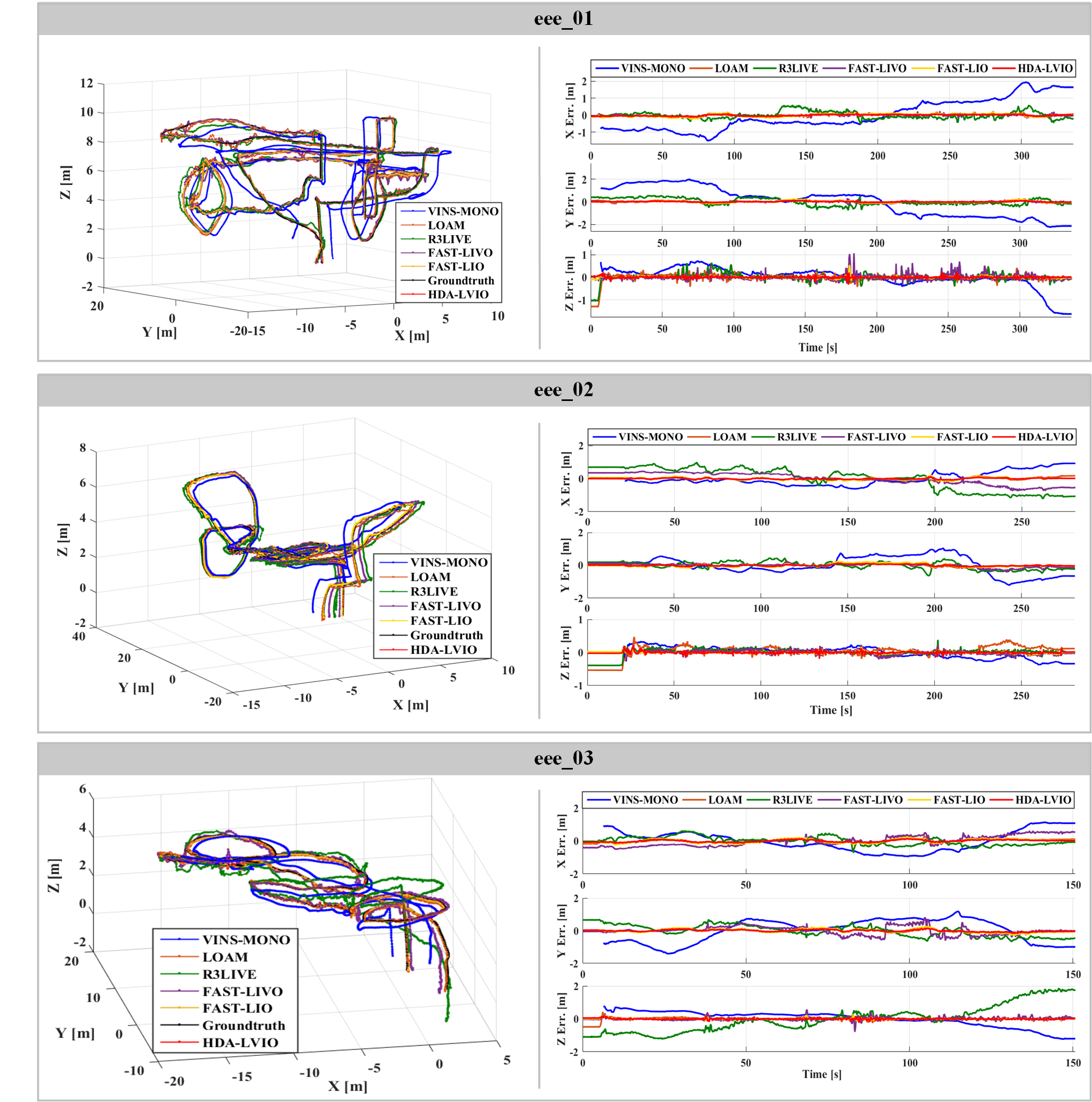}
	\caption{{The trajectory and localization error output from comparision algorithms and HDA-LVIO in segments \textbf{eee\_*} of NTU-VIRAL datasets.} }
	\label{viral eee}
\end{figure}
\begin{figure}[htbp]
	\centering
	\includegraphics[width=3.0in]{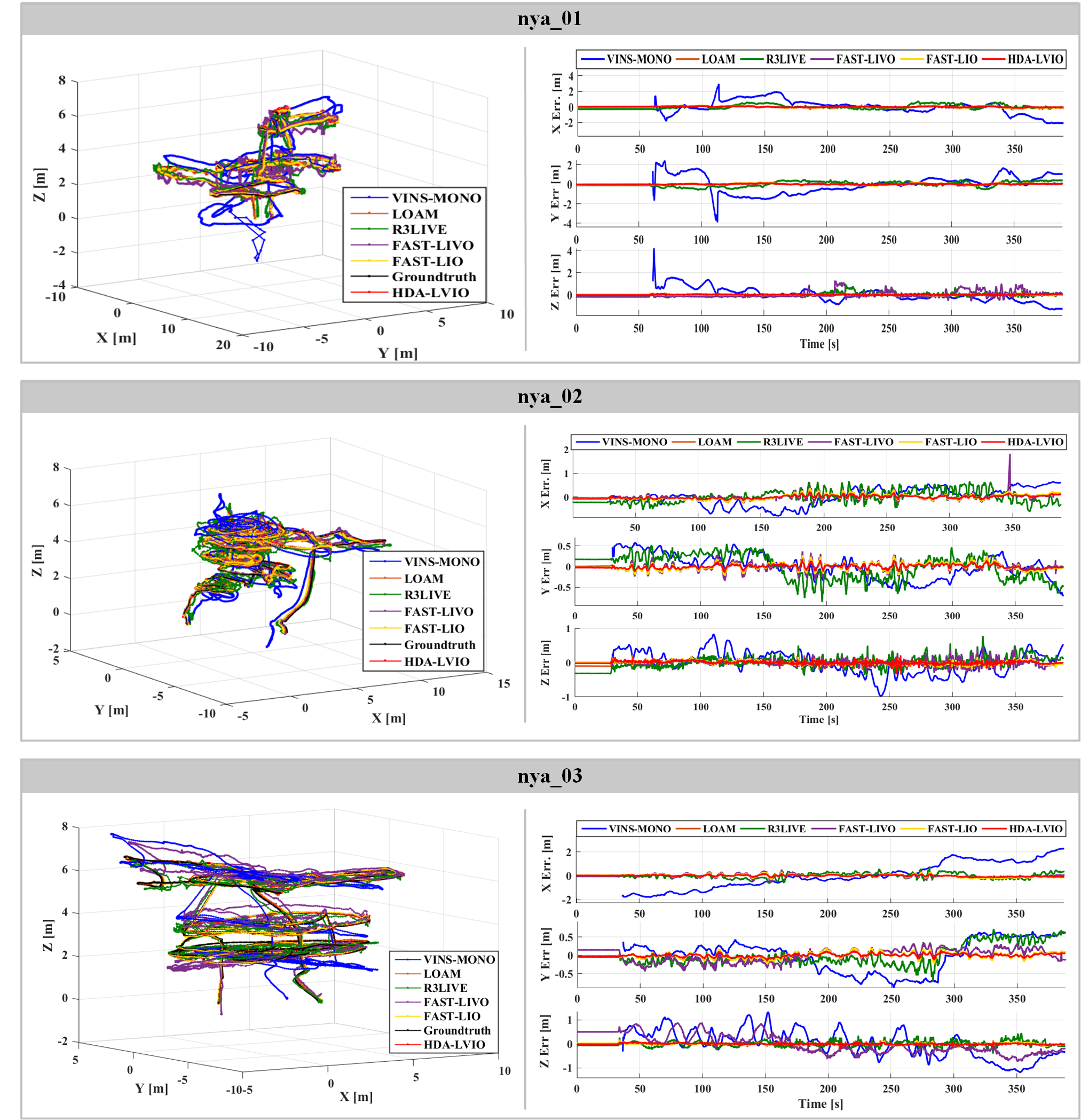}
	\caption{{The trajectory and localization error output from comparision algorithms and HDA-LVIO in segments \textbf{nya\_*} of NTU-VIRAL datasets.} }
	\label{viral nya}
\end{figure}
\begin{figure}[htbp]
	\centering
	\includegraphics[width=3.0in]{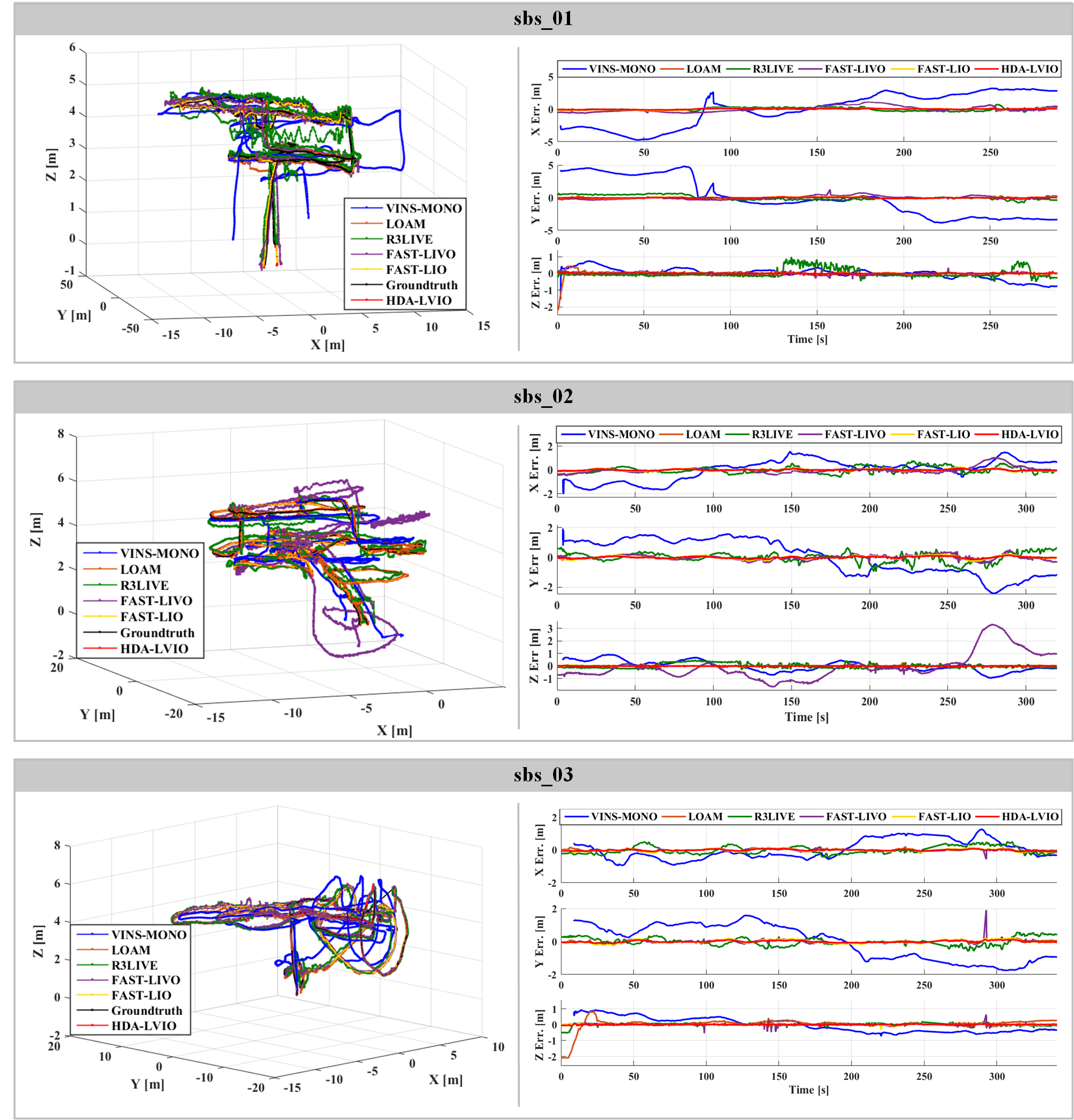}
	\caption{{The trajectory and localization error output from comparision algorithms and HDA-LVIO in segments \textbf{sbs\_*} of NTU-VIRAL datasets.}  }
	\label{viral sbs}
\end{figure}

\begin{table}[htbp]
	\centering
	\caption{{Runime of the comparison algorithms and the proposed HDA-LVIO in NTU-VIRAL dataset.}}
	\begingroup
	\setlength{\tabcolsep}{3pt} 
	\renewcommand{\arraystretch}{1.3} 
	\begin{tabular}{c|c|c|c|c|c|c}
		\Xhline{1.2px}
		\addlinespace[0.5ex]
		\multicolumn{1}{c|}{\textbf{\makecell[c]{runtime \\(ms)}}} & \multicolumn{1}{c|}{{\makecell[c]{VINS-\\MONO}}} & \multicolumn{1}{c|}{LOAM} & \multicolumn{1}{c|}{\makecell[c]{FAST-\\LIO}}& \multicolumn{1}{c|}{R3LIVE}& \multicolumn{1}{c|}{\makecell[c]{FAST-\\LIVO}}& \multicolumn{1}{c}{\makecell[c]{HDA-\\LVIO}} \\
		\Xhline{1.0px}
		LIS & -- & 38.24 & 25.75 & 23.54 & 28.37 & {34.79}  \\
		\hline
		VIS & 26.46 & -- & -- & 20.59 & 23.54 & {36.87} \\
		\addlinespace[0.5ex]
		\Xhline{1px}
	\end{tabular}%
	\endgroup
	\label{runtime}%
\end{table}%
In the NTU-VIRAL datasets, a DJI M600 Pro hexacopter is employed to carry the sensor setup. The sensor parameters are detailed in Table  \ref{NTU VIRAL SENSOR}.
We conduct comparative experiments using nine segments of data from this dataset, encompassing both indoor scenes (\textbf{nya}\_*) and outdoor scenes (\textbf{eee}\_* and \textbf{sbs}\_*). The localization results are presented in Fig. \ref{viral eee}, \ref{viral nya}, and \ref{viral sbs}. In detail, Fig. \ref{viral eee} illustrates the motion trajectory and localization error curves for the (\textbf{eee}\_*) data series, while Fig. \ref{viral nya} pertains to the (\textbf{nya}\_*) data series, and Fig. \ref{viral sbs} corresponds to the (\textbf{sbs}\_*) data series. Distinct colors are employed to differentiate trajectories and localization errors associated with different algorithms: VINS-MONO is represented by blue, LOAM by brown, R3LIVE by green, FAST-LIVO by purple, FAST-LIO by yellow, the proposed HDA-LVIO by red, and the groundtruth trajectory by black. The data segments selected from the NTU-VIRAL dataset encompass a significant amount of spatial motion, making them crucial for a comprehensive evaluation of the effectiveness of the proposed HDA-LVIO algorithm. 

The trajectory error plot clearly demonstrates that the proposed HDA-LVIO attains consistently minimal localization errors, virtually converging to zero meters  compared to other benchmark algorithms. To provide a qualitative assessment of the localization accuracy achieved by HDA-LVIO, we compute the ATE for each data segment, and the results are summarized in Table \ref{ate for viral}. The data presented in this table further emphasizes that the proposed algorithm consistently achieves the highest level of accuracy in localization on each data segment.

The improved localization accuracy of the proposed HDA-LVIO can primarily be attributed to three key factors. Firstly, the data segments selected from the NTU VIRAL dataset represent urban environments with numerous planar structures. And the proposed HDA-LVIO extracts the central point of these planes and projects onto the image, obtaining the projection point and establishing the reprojection error used for localization. Importantly, the depth of these planes undergo continuous and gradual changes. Even with a slight deviation in the initial state of image frame, the method ensures that the depth error of the projection points remains within an effective range. In contrast, both the R3LIVE and FAST-LIVO directly project LiDAR points onto the image. However, due to the presence of the edge scene in the environment, significant errors may be introduced into the depth of the projection points using these methods. These errors have a detrimental impact on the calculation accuracy of the reprojection error, ultimately leading to reduced localization accuracy.
Secondly, the proposed HDA-LVIO utilizes a sliding window approach to recover depth of image feature points, which relies on a sequence of accurate VIS state within the window. Furthermore, tracking failure of feature points resulting from rapid motion conditions are addressed based on the epipolar geometric constraints, ensuring thorough parallax between feature points within the sliding window to maintain depth recovery accuracy. 
Thirdly, in contrast to FAST-LIVO and R3LIVE, which solely rely  on projection points, the proposed HDA-LVIO leverages both projection  points and feature points to compute the reprojection error and employs them as observations for ESIKF to estimate VIS state. Since capitalizing on environmental information constraints from both LiDAR and Visual, the proposed method can enhance the estimation accuray of VIS state.

In addition to evaluating localization accuracy, we conduct a comparative analysis of the runtime between the proposed HDA-LVIO and other benchmark algorithms. Considering that the VIS and LIS operate concurrently within the proposed HDA-LVIO system, we compare their runntime individually. And the detail results are recorded in Table \ref{runtime}. To ensure a comprehensive comparison, even though LOAM is a LiDAR-based odometry without IMU assistance, we still evaluated its runntime performance alongside the LIS in the proposed HDA-LVIO. For the runtime of LIS, it is worth noting that the LIS implementations in R3LIVE and FAST-LIVO closely correspond  with FAST-LIO, and all three demonstrate comparable running time of approximately 25 ms. Conversely, the runntime of the LIS component in the proposed HDA-LVIO is measured at 34.79 ms. This difference can be attributed to the integration of an incremental adaptive plane extraction algorithm within the LIS component of the proposed HDA-LVIO. However, it is essential to emphasize that the proposed plane extraction algorithm employs an incremental strategy and utilizes a hash table to efficiently manage the extracted planes, ensuring that the runtime of LIS component in the proposed HDA-LVIO remains compliant with real-time operational requirements. 

Regarding the runtime of the VIS, the results in Table \ref{runtime} indicate that the VIS component in the proposed HDA-LVIO requires 36.87ms, which exceeds the runtime of other algorithms. This difference can be primarily attributed to two key reasons: 
Firstly, the VIS component in the proposed HDA-LVIO employs a sliding window approach for depth restoration of feature points. It also utilizes the epipolar geometric constraints-based method for recovering feature points that cannot be accurately tracked within the sliding window. 
Furthermore, the VIS component simultaneously employs both feature points and projection points to construct reprojection errors, which are used as observations within the ESIKF for state estimation. However, the runntime of VIS in the proposed HDA-LVIO shows that it meets the real-time requirements.


\subsection{KITTI datasets}
\begin{figure}[htbp]
	\centering
	\includegraphics[width=2.3in]{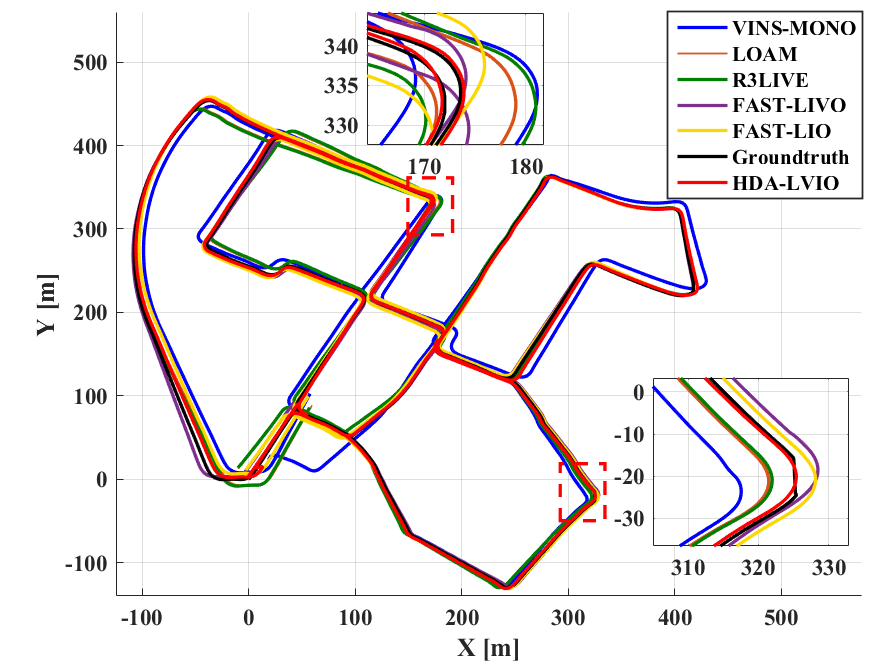}
	\caption{{The trajectory output from comparision algorithms and HDA-LVIO in  segment \textbf{00} of KITTI datasets.} }
	\label{traj_00}
\end{figure}

\begin{figure}[htbp]
	\centering
	\includegraphics[width=2.5in]{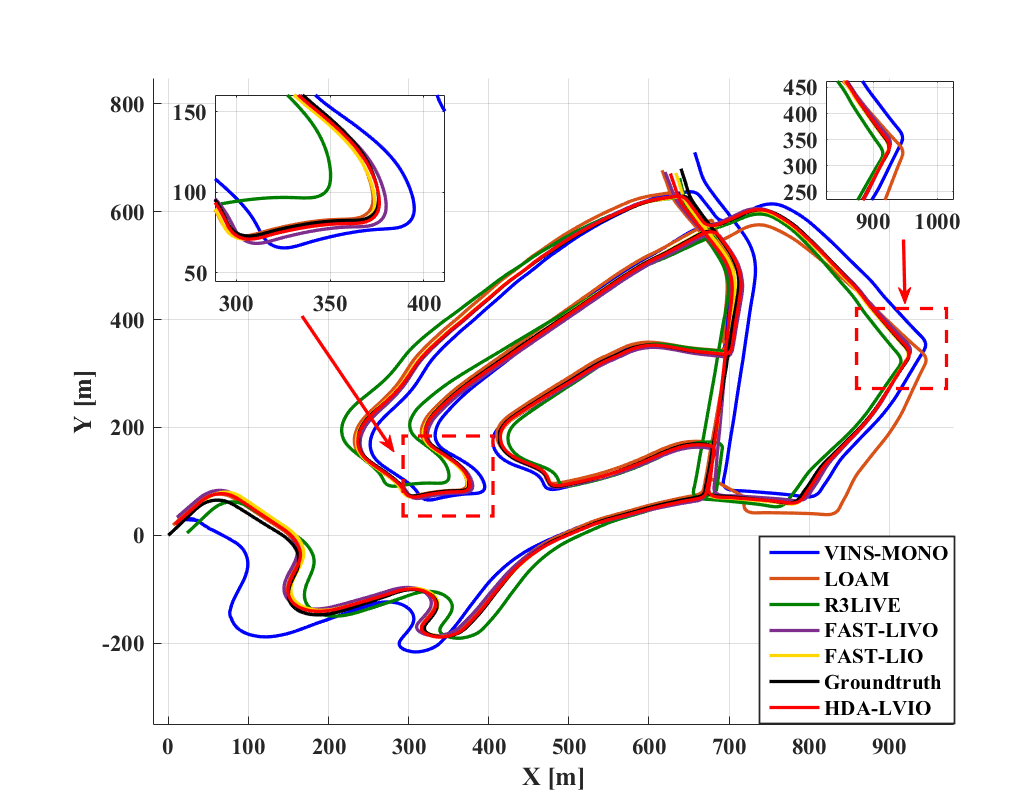}
	\caption{{The trajectory output from comparision algorithms and HDA-LVIO in  segment \textbf{02} of KITTI datasets.} }
	\label{traj_02}
\end{figure}

\begin{figure}[htbp]
	\centering
	\includegraphics[width=2.5in]{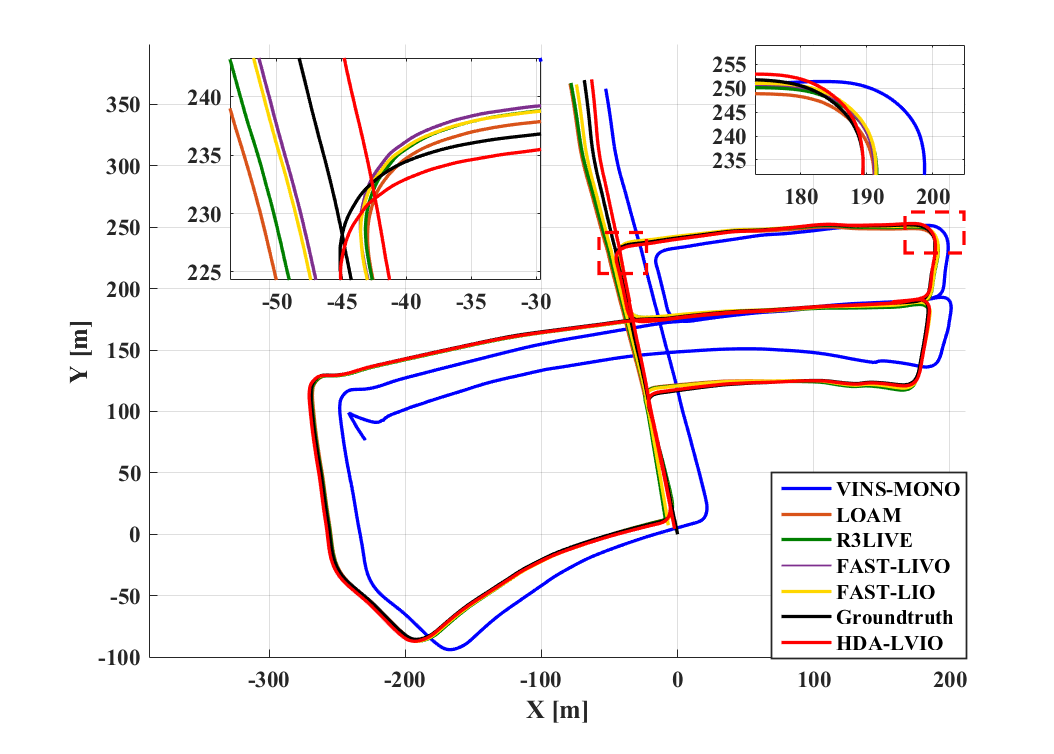}
	\caption{{The trajectory output from comparision algorithms and HDA-LVIO in  segment \textbf{05} of KITTI datasets.} }
	\label{traj_05}
\end{figure}

The KITTI dataset is a widely used SLAM dataset and the utilization of this dataset affords us the opportunity to juxtapose the proposed HDA-LVIO with a substantial corpus of preceding research. And the sensor parameters in KITTI are depicted in Table \ref{KITTI SENSOR}.
We chose data segments 00, 02, and 05 to validate the efficacy of the proposed HDA-LVIO. And the trajectory are depicted in Fig. \ref{traj_00}, \ref{traj_02} and \ref{traj_05}. The color of the trajectory in these figures are consistent with Sec. \ref{ntu-viral}. Considering that the KITTI dataset is gathered from a ground vehicle, we exclusively present trajectories within the X-Y plane. To effectively illustrate the enhancements in localization accuracy achieved by the proposed HDA-LVIO system, we magnify specific regions within the trajectory. These magnified regions are delineated by red dotted lines. These figures clearly demonstrate that the proposed algorithm yields significantly improvement in localization accuracy. For a more detailed quantitative analysis, we compute the ATE for these trajectories and depict the results in Table \ref{kitti ate}. The localization results clearly indicate that the proposed algorithm outperforms other algorithms in terms of localization accuracy.
\begin{table}[htbp]
	\centering
	\caption{The parameter of sensors utilized in KITTI datasets.}
	\renewcommand{\arraystretch}{1.3} 
	\begin{tabular}{>{\setlength{\parskip}{1ex}}cc}
		\toprule
		\textbf{Sensor} & \textbf{Description} \\
		\midrule
		\makecell[c]{Point Grey\\ Flea 2 \\ (FL2-14S3C-C)} & \makecell[c]{a color global-shutter \\ camera with 1.4 megapixels.}\\
		\addlinespace[0.5ex]
		\hline
		\addlinespace[0.5ex]
		\makecell[c]{OXTS TR\\ 3003} &\makecell[c]{ an integrated navigation system,\\ utilizing MEMS IMU (100Hz) and \\ RTK,achieves a localization error of \\ less than 5 cm, serving as\\ the ground truth for localization.}\\
		\addlinespace[0.5ex]
		\hline
		\addlinespace[0.5ex]
		\makecell[c]{Velodyne\\ HDL-64E}  & \makecell[c]{a 64-channel ($-24.33^{\circ}, 2^{\circ}$) LiDAR,\\ rotating at 10 Hz. And the\\ azimuth angular resolution is $0.09^{\circ}$. }  \\
		\bottomrule
	\end{tabular}%
	\label{KITTI SENSOR}%
\end{table}%

\begin{table*}[htbp]
	\centering
	\caption{{ATE of the comparison algorithms and the proposed HDA-LVIO in KITTI dataset.}}
	\begingroup
	\setlength{\tabcolsep}{6pt} 
	\renewcommand{\arraystretch}{1.3} 
	\begin{tabular}{c|c|c|c|c|c|c}
		\Xhline{1.2px}
		\multicolumn{1}{c|}{\textbf{ATE (m)}} & \multicolumn{1}{c|}{{VINS-MONO}} & \multicolumn{1}{c|}{LOAM} & \multicolumn{1}{c|}{FAST-LIO}& \multicolumn{1}{c|}{R3LIVE}& \multicolumn{1}{c|}{FAST-LIVO}& \multicolumn{1}{c}{HDA-LVIO} \\
		\Xhline{1.0px}
		00 & 14.97 & 8.24 & 5.75 & 8.27 & 6.13 & \textbf{2.94}  \\
		\hline
		02 & 26.46 & 17.73 & 7.22 & 14.57 & 9.32 & \textbf{5.14} \\
		\hline
		05 & 19.35 & 4.29 & 3.71 & 3.69 & 3.16 & \textbf{2.03}  \\
		\Xhline{1px}
	\end{tabular}%
	\endgroup
	\label{kitti ate}%
\end{table*}%

\subsection{Our platform data}
\begin{figure}[htbp]
	\centering
	\includegraphics[width=2.5in]{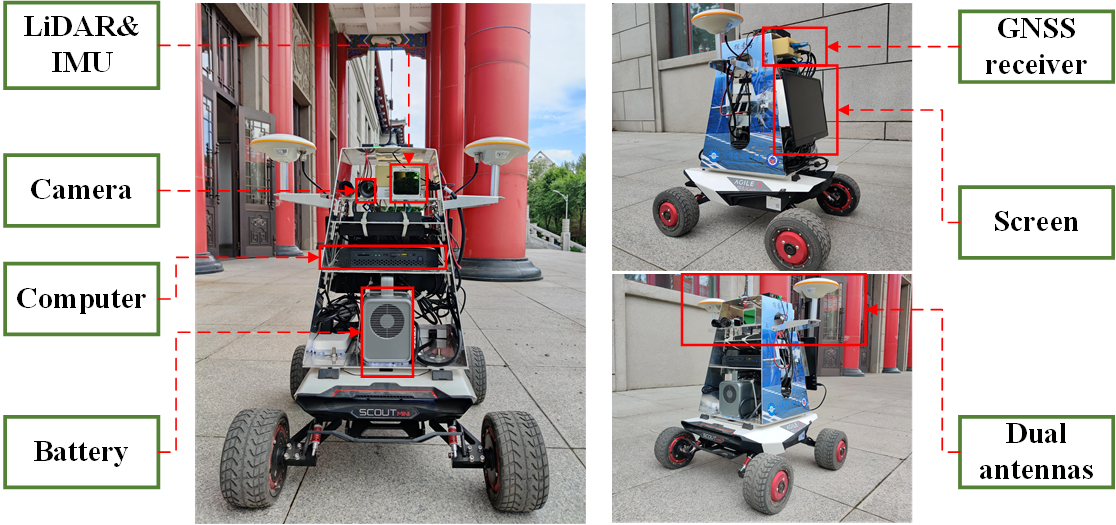}
	\caption{{The sensor suite utilized in our platform.} }
	\label{equipment}
\end{figure}

\begin{table}[htbp]
	\centering
	\caption{The parameter of sensors utilized in our platform.}
	\renewcommand{\arraystretch}{1.3} 
	\begin{tabular}{cc}
		\toprule
		\textbf{Sensor} & \textbf{Description} \\
		\midrule
		\makecell[c]{Basler \\ acA1440-73gc} & \makecell[c]{delivers 20 frames per\\ second  at 1.6 MP resolution. }\\
		\addlinespace[0.5ex]
		\hline
		\addlinespace[0.5ex]
		\makecell[c]{BMI088} &\makecell[c]{ a MEMS IMU rating at 200 Hz,\\ built-in the Livox Avia.}\\
		\addlinespace[0.5ex]
		\hline
		\addlinespace[0.5ex]
		\makecell[c]{Livox Avia}  & \makecell[c]{a non-repetitive scanning LiDAR with \\the horizontal field of view (FOV) \\ of $70.4^{\circ}$ and vertical FOV of $77.2^{\circ}$.}  \\
		\addlinespace[0.5ex]
		\hline
		\addlinespace[0.5ex]
		\makecell[c]{RTK} & \makecell[c]{with localization error less\\  than 5 cm,is adopted as \\the ground truth for localization.  } \\
		\addlinespace[0.5ex]
		\hline
		\addlinespace[0.5ex]
		\makecell[c]{Scout mini} & \makecell[c]{ a differential steering chassis with \\ speed up to $10.8$ km/h and \\ angular velocity up to $150 ^ \circ / s$.} \\
		\bottomrule
	\end{tabular}%
	\label{our platform SENSOR}%
\end{table}%
To further validate the improvement of the proposed HDA-LVIO system in localization, an experimental platform is established as Fig. \ref{equipment} and the parameter of the sensor suite are depicted in Table \ref{our platform SENSOR}.
We collected data both in indoors and outdoors, where outdoor data used RTK as a reference, and indoor accuracy is determined based on position deviation between the starting and ending points. Due to significant drift observed in VINS-MONO, possibly due to limited vehicle height to capture adequate feature information, we opted not to compare it with the proposed HDA-LVIO.

\subsubsection{Outdoor data}

\begin{figure}[htbp]
	\centering
	\includegraphics[width=2.0in]{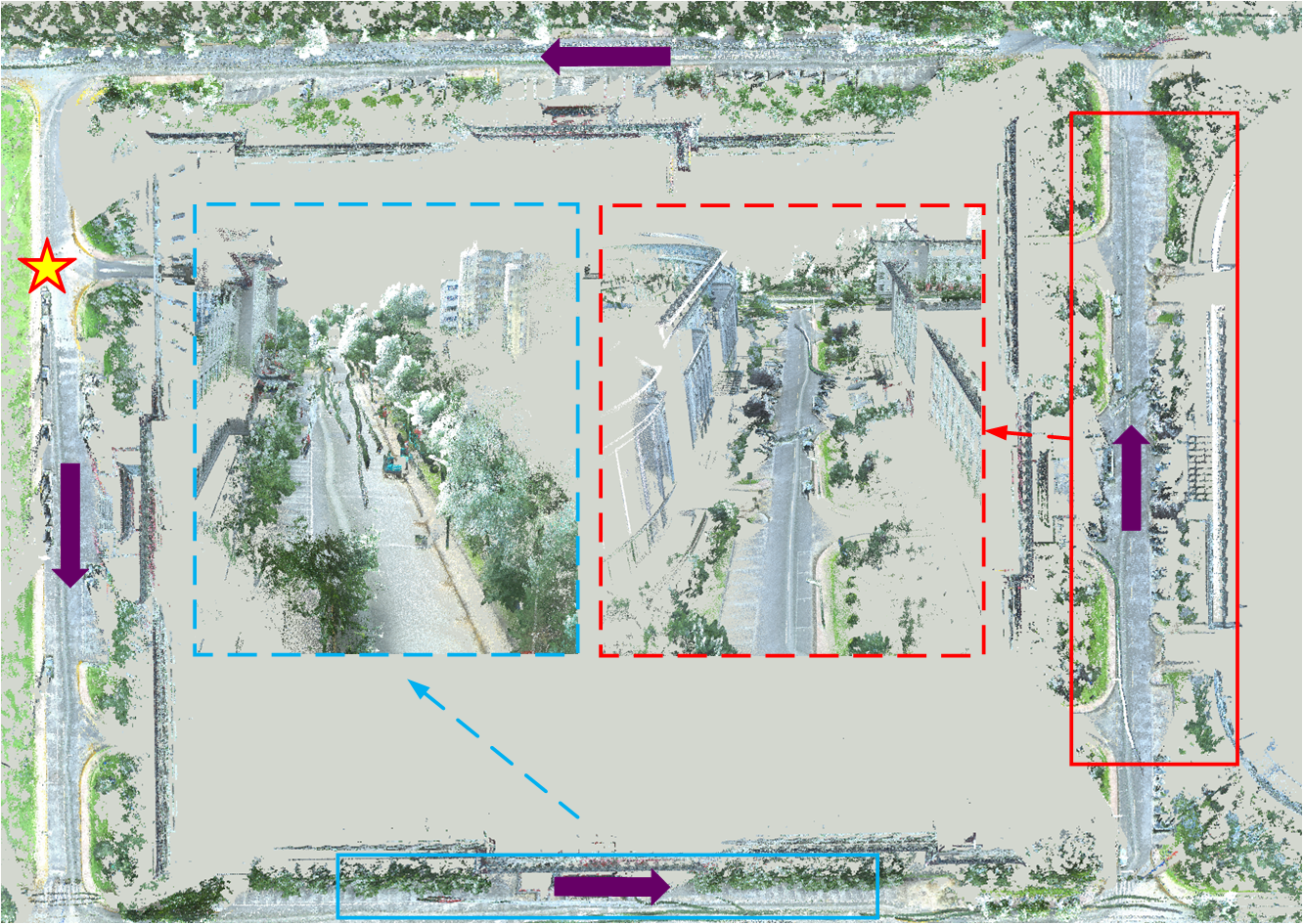}
	\caption{{The outdoor map established with the proposed HDA-LVIO.} }
	\label{outdoor_map}
\end{figure}

We gathered environmental data around the teaching building to validate the effectiveness of our proposed algorithm. The map generated by our HDA-LVIO approach is dipicted in Fig. \ref{outdoor_map}, with the yellow star representing the initial position and the purple arrow indicating the general motion direction. To facilitate a clearly examination of the map's details, we zoom in on two specific regions, respectively represented by the blue and red rectangles.

\begin{figure}[htbp]
	\centering
	\includegraphics[width=2.5in]{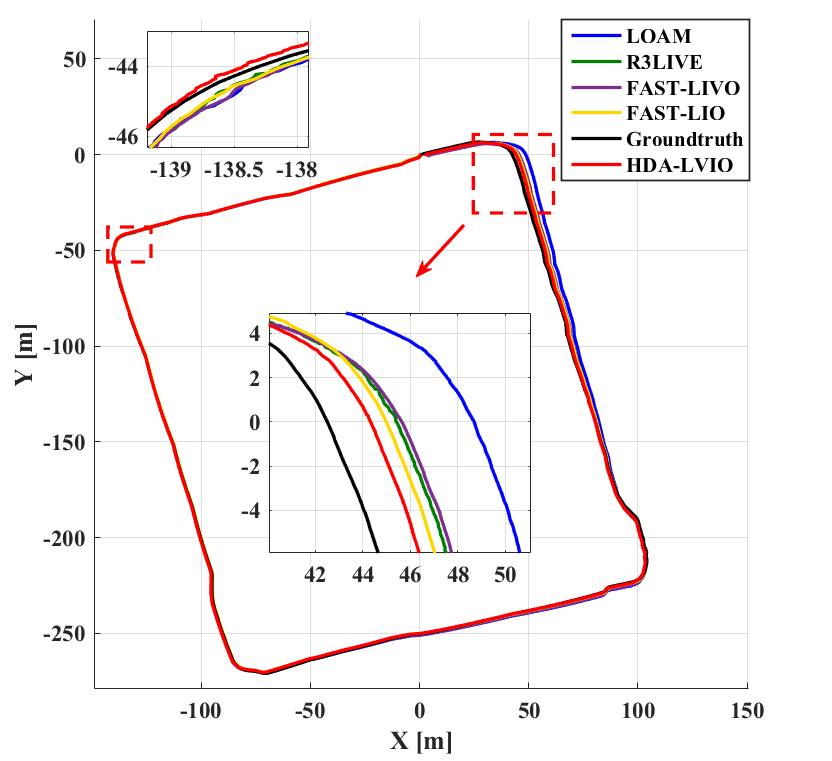}
	\caption{{The localization result with our platform in outdoor environment.} }
	\label{our_outdoor}
\end{figure}

The localization results of the reference algorithm and the proposed algorithm are shown in Fig. \ref{our_outdoor}. We zoom in on two specific areas and outlined them with red rectangles for closer examination. Upon closer inspection of the enlarged image, it becomes evident that the localization result generated by the proposed HDA-LVIO closely aligns with the ground truth trajectory.

\begin{table}[htbp]
	\centering
	\caption{{ATE of the comparison algorithms and the proposed HDA-LVIO in data from our platform.}}
	\begingroup
	\setlength{\tabcolsep}{5pt} 
	\renewcommand{\arraystretch}{1.3} 
	\begin{tabular}{c|c|c|c|c|c}
		\Xhline{1.2px}
		\addlinespace[0.5ex]
		\multicolumn{1}{c|}{\textbf{ATE (m)}} & \multicolumn{1}{c|}{LOAM} & \multicolumn{1}{c|}{\makecell[c]{FAST-\\LIO}}& \multicolumn{1}{c|}{R3LIVE}& \multicolumn{1}{c|}{\makecell[c]{FAST-\\LIVO}}& \multicolumn{1}{c}{\makecell[c]{HDA-\\LVIO}} \\
		\Xhline{1.0px}
		outdoor  & {8.57} & {2.19} & {2.63} & {3.74} & \textbf{1.56} \\
		\hline
		indoor  & 31.08 & 12.63 & 14.77 & 24.70 & \textbf{7.32}  \\
		\addlinespace[0.5ex]
		\Xhline{1px}
	\end{tabular}%
	\endgroup
	\label{ourplatform}%
\end{table}

\subsubsection{Indoor data}

\begin{figure}[htbp]
	\centering
	\includegraphics[width=2.0in]{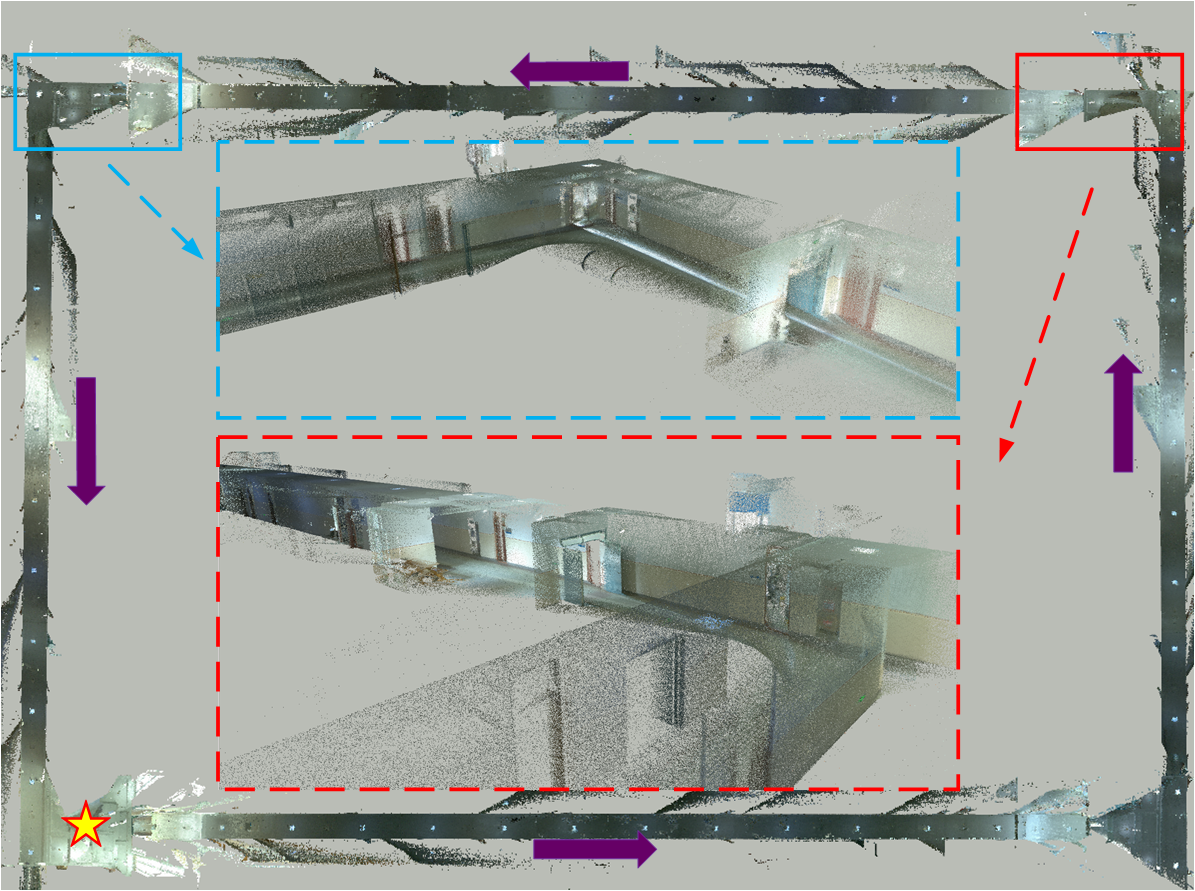}
	\caption{{The indoor map established with the proposed HDA-LVIO.} }
	\label{indoor_map}
\end{figure}

We collected indoor data within the teaching building, and to better visualize positioning drift, we devised a circular route while employing ground markers to ensure the distance between the start and end points remained within 1cm. The indoor environment map constructed with the proposed algorithm is shown in Fig. \ref{indoor_map}, where the yellow star points represent the starting and ending points of the trajectory, and the purple arrows are the approximate running direction of the trajectory. We have chosen to zoom in on two specific areas to provide a clearer illustration of the precision of our map construction. As depicted in the enlarged images, the map generated by our proposed HDA-LVIO method realistically and vividly represents the environmental details. Notably, our testing took place in a challenging corridor-like environment, where mapping and positioning posed greater difficulties. However, our proposed algorithm, leveraging both planar data from LiDAR and feature information from images, has demonstrated its capability to yield favorable mapping outcomes even in such demanding environments.

\begin{figure}[H]
	\centering
	\includegraphics[width=2.5in]{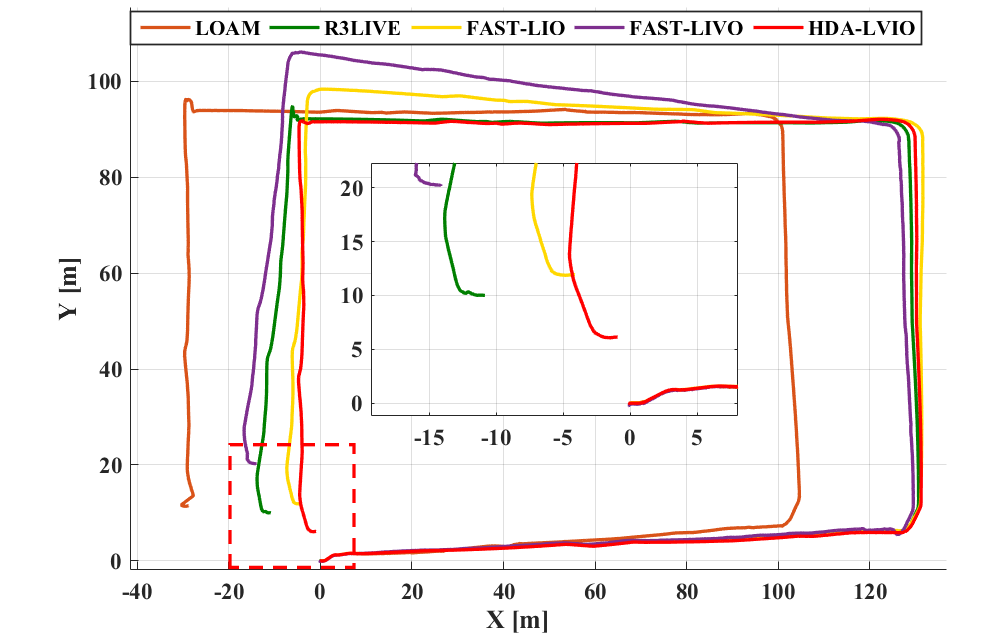}
	\caption{{The localization result with our platform in indoor environment.} }
	\label{our_indoor}
\end{figure}

The trajectories obtained by the proposed algorithm and the comparison algorithm are shown in Fig.\ref{our_indoor}. As depicted in the figure, the trajectory produced by the proposed HDA-LVIO closely resembles a rectangle, aligning well with the actual environment. Furthermore, the end point of the trajectory generated by the proposed method is in proximity to the start point, providing qualitative validation of the algorithm's effectiveness in localization. Simultaneously, we calculate the distance between the starting point and the end point, and these values are recorded in Table \ref{ourplatform}. The distance between the starting and final points for the proposed algorithms is 7.32 meters, whereas other algorithms exceed 10 meters in distance. The most substantial deviation is observed with LOAM, which extends up to 30 meters. This observation underscores the capacity of the proposed algorithm to significantly enhance localization accuracy, even in challenging scenarios, such as corridors.

\section{Conclusion}
HDA-LVIO, a LiDAR-Visual-Inertial odometry system based on hybrid data association, is introduced with the primary goal of achieving high-precision localization and mapping in urban environments. HDA-LVIO comprises two subsystems: VIS and LIS. Within the LIS component, we employ an ESIKF for global map construction. Furthermore, an innovative incremental adaptive plane extraction algorithm is incorporated for environmental plane extraction from the global map constructed by LIS.
In the VIS component, sliding window optimization is adopted for feature point depth estimation and  the epipolar geometric constraints-based approach is employed to recover feature points that may fail to be tracked within the sliding window. Subsequently, both projection points and feature points are concurrently employed to formulate reprojection errors, which serve as observation values for ESIKF in the VIS.
Extensive experiments are conducted to validate the accurate localization capabilities of HDA-LVIO within urban scenarios. Nonetheless, two key issues have been identified during these experiments. Firstly, a challenge is posed by the increased computational load of the HDA-LVIO algorithm, attributed to its utilization of both projection and feature points for VIS state estimation, coupled with the inclusion of plane extraction and feature point estimation algorithms. Additionally, the presence of dynamic objects in the urban environment may impact the hybrid data association in the proposed HDA-LVIO system and potentially result in reduced accuracy. Therefore, our future research efforts will focus on mitigating these challenges by reducing the computational complexity of the HDA-LVIO algorithm and addressing the influence of dynamic objects.

\bibliographystyle{IEEEtran}
\bibliography{ieee}

\begin{thebibliography}{10}
\providecommand{\url}[1]{#1}
\csname url@samestyle\endcsname
\providecommand{\newblock}{\relax}
\providecommand{\bibinfo}[2]{#2}
\providecommand{\BIBentrySTDinterwordspacing}{\spaceskip=0pt\relax}
\providecommand{\BIBentryALTinterwordstretchfactor}{4}
\providecommand{\BIBentryALTinterwordspacing}{\spaceskip=\fontdimen2\font plus
\BIBentryALTinterwordstretchfactor\fontdimen3\font minus
  \fontdimen4\font\relax}
\providecommand{\BIBforeignlanguage}[2]{{%
\expandafter\ifx\csname l@#1\endcsname\relax
\typeout{** WARNING: IEEEtran.bst: No hyphenation pattern has been}%
\typeout{** loaded for the language `#1'. Using the pattern for}%
\typeout{** the default language instead.}%
\else
\language=\csname l@#1\endcsname
\fi
#2}}
\providecommand{\BIBdecl}{\relax}
\BIBdecl

\bibitem{kuutti2018survey}
S.~Kuutti, S.~Fallah, K.~Katsaros, M.~Dianati, F.~Mccullough, and
  A.~Mouzakitis, ``A survey of the state-of-the-art localization techniques and
  their potentials for autonomous vehicle applications,'' \emph{IEEE Internet
  of Things Journal}, vol.~5, no.~2, pp. 829--846, 2018.

\bibitem{cadena2016past}
C.~Cadena, L.~Carlone, H.~Carrillo, Y.~Latif, D.~Scaramuzza, J.~Neira, I.~Reid,
  and J.~J. Leonard, ``Past, present, and future of simultaneous localization
  and mapping: Toward the robust-perception age,'' \emph{IEEE Transactions on
  robotics}, vol.~32, no.~6, pp. 1309--1332, 2016.

\bibitem{zhang2014loam}
J.~Zhang and S.~Singh, ``Loam: Lidar odometry and mapping in real-time.'' in
  \emph{Robotics: Science and systems}, vol.~2, no.~9.\hskip 1em plus 0.5em
  minus 0.4em\relax Berkeley, CA, 2014, pp. 1--9.

\bibitem{ASL-SLAM}
B.~Zhou, C.~Li, S.~Chen, D.~Xie, M.~Yu, and Q.~Li, ``Asl-slam: A lidar slam
  with activity semantics-based loop closure,'' \emph{IEEE Sensors Journal},
  vol.~23, no.~12, pp. 13\,499--13\,510, 2023.

\bibitem{WangZhoubo}
Z.~Wang, Z.~Zhang, X.~Kang, M.~Wu, S.~Chen, and Q.~Li, ``Dor-lins: Dynamic
  objects removal lidar-inertial slam based on ground pseudo occupancy,''
  \emph{IEEE Sensors Journal}, vol.~23, no.~20, pp. 24\,907--24\,915, 2023.

\bibitem{qin2018vins}
T.~Qin, P.~Li, and S.~Shen, ``Vins-mono: A robust and versatile monocular
  visual-inertial state estimator,'' \emph{IEEE Transactions on Robotics},
  vol.~34, no.~4, pp. 1004--1020, 2018.

\bibitem{SunTian}
T.~Sun, Y.~Liu, Y.~Wang, and Z.~Xiao, ``An improved monocular visual-inertial
  navigation system,'' \emph{IEEE Sensors Journal}, vol.~21, no.~10, pp.
  11\,728--11\,739, 2021.

\bibitem{Chou}
C.-C. Chou and C.-F. Chou, ``Efficient and accurate tightly-coupled
  visual-lidar slam,'' \emph{IEEE Transactions on Intelligent Transportation
  Systems}, vol.~23, no.~9, pp. 14\,509--14\,523, 2022.

\bibitem{shan2018lego}
T.~Shan and B.~Englot, ``Lego-loam: Lightweight and ground-optimized lidar
  odometry and mapping on variable terrain,'' in \emph{2018 IEEE/RSJ
  International Conference on Intelligent Robots and Systems (IROS)}.\hskip 1em
  plus 0.5em minus 0.4em\relax IEEE, 2018, pp. 4758--4765.

\bibitem{park2017illumination}
S.~Park, T.~Sch{\"o}ps, and M.~Pollefeys, ``Illumination change robustness in
  direct visual slam,'' in \emph{2017 IEEE international conference on robotics
  and automation (ICRA)}.\hskip 1em plus 0.5em minus 0.4em\relax IEEE, 2017,
  pp. 4523--4530.

\bibitem{strasdat2010scale}
H.~Strasdat, J.~Montiel, and A.~J. Davison, ``Scale drift-aware large scale
  monocular slam,'' \emph{Robotics: science and Systems VI}, vol.~2, no.~3,
  p.~7, 2010.

\bibitem{wang2019robust}
Z.~Wang, J.~Zhang, S.~Chen, C.~Yuan, J.~Zhang, and J.~Zhang, ``Robust high
  accuracy visual-inertial-laser slam system,'' in \emph{2019 IEEE/RSJ
  International Conference on Intelligent Robots and Systems (IROS)}.\hskip 1em
  plus 0.5em minus 0.4em\relax IEEE, 2019, pp. 6636--6641.

\bibitem{zuo2019lic}
X.~Zuo, P.~Geneva, W.~Lee, Y.~Liu, and G.~Huang, ``Lic-fusion:
  Lidar-inertial-camera odometry,'' in \emph{2019 IEEE/RSJ International
  Conference on Intelligent Robots and Systems (IROS)}.\hskip 1em plus 0.5em
  minus 0.4em\relax IEEE, 2019, pp. 5848--5854.

\bibitem{zuo2020lic}
X.~Zuo, Y.~Yang, P.~Geneva, J.~Lv, Y.~Liu, G.~Huang, and M.~Pollefeys,
  ``Lic-fusion 2.0: Lidar-inertial-camera odometry with sliding-window
  plane-feature tracking,'' in \emph{2020 IEEE/RSJ International Conference on
  Intelligent Robots and Systems (IROS)}.\hskip 1em plus 0.5em minus
  0.4em\relax IEEE, 2020, pp. 5112--5119.

\bibitem{zhao2021super}
S.~Zhao, H.~Zhang, P.~Wang, L.~Nogueira, and S.~Scherer, ``Super odometry:
  Imu-centric lidar-visual-inertial estimator for challenging environments,''
  in \emph{2021 IEEE/RSJ International Conference on Intelligent Robots and
  Systems (IROS)}.\hskip 1em plus 0.5em minus 0.4em\relax IEEE, 2021, pp.
  8729--8736.

\bibitem{jia2021lvio}
Y.~Jia, H.~Luo, F.~Zhao, G.~Jiang, Y.~Li, J.~Yan, Z.~Jiang, and Z.~Wang,
  ``Lvio-fusion: A self-adaptive multi-sensor fusion slam framework using
  actor-critic method,'' in \emph{2021 IEEE/RSJ International Conference on
  Intelligent Robots and Systems (IROS)}.\hskip 1em plus 0.5em minus
  0.4em\relax IEEE, 2021, pp. 286--293.

\bibitem{gao2022vido}
Y.~Gao, J.~Yuan, J.~Jiang, Q.~Sun, and X.~Zhang, ``Vido: A robust and
  consistent monocular visual-inertial-depth odometry,'' \emph{IEEE
  Transactions on Intelligent Transportation Systems}, vol.~24, no.~3, pp.
  2976--2992, 2022.

\bibitem{huang2018joint}
K.~Huang and C.~Stachniss, ``Joint ego-motion estimation using a laser scanner
  and a monocular camera through relative orientation estimation and 1-dof
  icp,'' in \emph{2018 IEEE/RSJ International Conference on Intelligent Robots
  and Systems (IROS)}.\hskip 1em plus 0.5em minus 0.4em\relax IEEE, 2018, pp.
  671--677.

\bibitem{zhang2017real}
J.~Zhang, M.~Kaess, and S.~Singh, ``A real-time method for depth enhanced
  visual odometry,'' \emph{Autonomous Robots}, vol.~41, pp. 31--43, 2017.

\bibitem{zhang2018laser}
J.~Zhang and S.~Singh, ``Laser--visual--inertial odometry and mapping with high
  robustness and low drift,'' \emph{Journal of field robotics}, vol.~35, no.~8,
  pp. 1242--1264, 2018.

\bibitem{shan2021lvi}
T.~Shan, B.~Englot, C.~Ratti, and D.~Rus, ``Lvi-sam: Tightly-coupled
  lidar-visual-inertial odometry via smoothing and mapping,'' in \emph{2021
  IEEE international conference on robotics and automation (ICRA)}.\hskip 1em
  plus 0.5em minus 0.4em\relax IEEE, 2021, pp. 5692--5698.

\bibitem{wang2022mvil}
Y.~Wang and H.~Ma, ``mvil-fusion: Monocular visual-inertial-lidar simultaneous
  localization and mapping in challenging environments,'' \emph{IEEE Robotics
  and Automation Letters}, vol.~8, no.~2, pp. 504--511, 2022.

\bibitem{shu2022multi}
C.~Shu and Y.~Luo, ``Multi-modal feature constraint based tightly coupled
  monocular visual-lidar odometry and mapping,'' \emph{IEEE Transactions on
  Intelligent Vehicles}, 2022.

\bibitem{tang2023vins}
H.~Tang, X.~Niu, T.~Zhang, L.~Wang, and J.~Liu, ``Le-vins: A robust
  solid-state-lidar-enhanced visual-inertial navigation system for low-speed
  robots,'' \emph{IEEE Transactions on Instrumentation and Measurement},
  vol.~72, pp. 1--13, 2023.

\bibitem{lowe2018complementary}
T.~Lowe, S.~Kim, and M.~Cox, ``Complementary perception for handheld slam,''
  \emph{IEEE Robotics and Automation Letters}, vol.~3, no.~2, pp. 1104--1111,
  2018.

\bibitem{zhu2021camvox}
Y.~Zhu, C.~Zheng, C.~Yuan, X.~Huang, and X.~Hong, ``Camvox: A low-cost and
  accurate lidar-assisted visual slam system,'' in \emph{2021 IEEE
  International Conference on Robotics and Automation (ICRA)}.\hskip 1em plus
  0.5em minus 0.4em\relax IEEE, 2021, pp. 5049--5055.

\bibitem{yin2022novel}
J.~Yin, D.~Luo, F.~Yan, and Y.~Zhuang, ``A novel lidar-assisted monocular
  visual slam framework for mobile robots in outdoor environments,'' \emph{IEEE
  Transactions on Instrumentation and Measurement}, vol.~71, pp. 1--11, 2022.

\bibitem{wang2021vanishing}
P.~Wang, Z.~Fang, S.~Zhao, Y.~Chen, M.~Zhou, and S.~An, ``Vanishing point aided
  lidar-visual-inertial estimator,'' in \emph{2021 IEEE International
  Conference on Robotics and Automation (ICRA)}.\hskip 1em plus 0.5em minus
  0.4em\relax IEEE, 2021, pp. 13\,120--13\,126.

\bibitem{giubilato2018scale}
R.~Giubilato, S.~Chiodini, M.~Pertile, and S.~Debei, ``Scale correct monocular
  visual odometry using a lidar altimeter,'' in \emph{2018 IEEE/RSJ
  international conference on intelligent robots and systems (IROS)}.\hskip 1em
  plus 0.5em minus 0.4em\relax IEEE, 2018, pp. 3694--3700.

\bibitem{graeter2018limo}
J.~Graeter, A.~Wilczynski, and M.~Lauer, ``Limo: Lidar-monocular visual
  odometry,'' in \emph{2018 IEEE/RSJ international conference on intelligent
  robots and systems (IROS)}.\hskip 1em plus 0.5em minus 0.4em\relax IEEE,
  2018, pp. 7872--7879.

\bibitem{shin2018direct}
Y.-S. Shin, Y.~S. Park, and A.~Kim, ``Direct visual slam using sparse depth for
  camera-lidar system,'' in \emph{2018 IEEE International Conference on
  Robotics and Automation (ICRA)}.\hskip 1em plus 0.5em minus 0.4em\relax IEEE,
  2018, pp. 5144--5151.

\bibitem{shin2020dvl}
Y.-S. Shin, Y.~S. Park, and A.~Kim, ``Dvl-slam: Sparse depth enhanced direct
  visual-lidar slam,'' \emph{Autonomous Robots}, vol.~44, no.~2, pp. 115--130,
  2020.

\bibitem{huang2019accurate}
K.~Huang, J.~Xiao, and C.~Stachniss, ``Accurate direct visual-laser odometry
  with explicit occlusion handling and plane detection,'' in \emph{2019
  International Conference on Robotics and Automation (ICRA)}.\hskip 1em plus
  0.5em minus 0.4em\relax IEEE, 2019, pp. 1295--1301.

\bibitem{huang2020lidar}
S.-S. Huang, Z.-Y. Ma, T.-J. Mu, H.~Fu, and S.-M. Hu, ``Lidar-monocular visual
  odometry using point and line features,'' in \emph{2020 IEEE International
  Conference on Robotics and Automation (ICRA)}.\hskip 1em plus 0.5em minus
  0.4em\relax IEEE, 2020, pp. 1091--1097.

\bibitem{zheng2022fast}
C.~Zheng, Q.~Zhu, W.~Xu, X.~Liu, Q.~Guo, and F.~Zhang, ``Fast-livo: Fast and
  tightly-coupled sparse-direct lidar-inertial-visual odometry,'' in \emph{2022
  IEEE/RSJ International Conference on Intelligent Robots and Systems
  (IROS)}.\hskip 1em plus 0.5em minus 0.4em\relax IEEE, 2022, pp. 4003--4009.

\bibitem{lin2022r}
J.~Lin and F.~Zhang, ``R 3 live: A robust, real-time, rgb-colored,
  lidar-inertial-visual tightly-coupled state estimation and mapping package,''
  in \emph{2022 International Conference on Robotics and Automation
  (ICRA)}.\hskip 1em plus 0.5em minus 0.4em\relax IEEE, 2022, pp.
  10\,672--10\,678.

\bibitem{yuan2023sdv}
Z.~Yuan, Q.~Wang, K.~Cheng, T.~Hao, and X.~Yang, ``Sdv-loam: Semi-direct
  visual-lidar odometry and mapping,'' \emph{IEEE Transactions on Pattern
  Analysis and Machine Intelligence}, 2023.

\bibitem{harris1988combined}
C.~Harris, M.~Stephens \emph{et~al.}, ``A combined corner and edge detector,''
  in \emph{Alvey vision conference}, vol.~15, no.~50.\hskip 1em plus 0.5em
  minus 0.4em\relax Citeseer, 1988, pp. 10--5244.

\bibitem{kaess2012isam2}
M.~Kaess, H.~Johannsson, R.~Roberts, V.~Ila, J.~J. Leonard, and F.~Dellaert,
  ``isam2: Incremental smoothing and mapping using the bayes tree,'' \emph{The
  International Journal of Robotics Research}, vol.~31, no.~2, pp. 216--235,
  2012.

\bibitem{xu2022fast}
W.~Xu, Y.~Cai, D.~He, J.~Lin, and F.~Zhang, ``Fast-lio2: Fast direct
  lidar-inertial odometry,'' \emph{IEEE Transactions on Robotics}, vol.~38,
  no.~4, pp. 2053--2073, 2022.

\bibitem{adelson1984pyramid}
E.~H. Adelson, C.~H. Anderson, J.~R. Bergen, P.~J. Burt, and J.~M. Ogden,
  ``Pyramid methods in image processing,'' \emph{RCA engineer}, vol.~29, no.~6,
  pp. 33--41, 1984.

\bibitem{lucas1981iterative}
B.~D. Lucas and T.~Kanade, ``An iterative image registration technique with an
  application to stereo vision,'' in \emph{IJCAI'81: 7th international joint
  conference on Artificial intelligence}, vol.~2, 1981, pp. 674--679.

\bibitem{more2006levenberg}
J.~J. Mor{\'e}, ``The levenberg-marquardt algorithm: implementation and
  theory,'' in \emph{Numerical analysis: proceedings of the biennial Conference
  held at Dundee, June 28--July 1, 1977}.\hskip 1em plus 0.5em minus
  0.4em\relax Springer, 2006, pp. 105--116.

\bibitem{he2021kalman}
D.~He, W.~Xu, and F.~Zhang, ``Kalman filters on differentiable manifolds,''
  \emph{arXiv preprint arXiv:2102.03804}, 2021.

\bibitem{chum2003locally}
O.~Chum, J.~Matas, and J.~Kittler, ``Locally optimized ransac,'' in
  \emph{Pattern Recognition: 25th DAGM Symposium, Magdeburg, Germany, September
  10-12, 2003. Proceedings 25}.\hskip 1em plus 0.5em minus 0.4em\relax
  Springer, 2003, pp. 236--243.

\bibitem{lin2021r}
J.~Lin, C.~Zheng, W.~Xu, and F.~Zhang, ``R2live: A robust, real-time,
  lidar-inertial-visual tightly-coupled state estimator and mapping,''
  \emph{IEEE Robotics and Automation Letters}, vol.~6, no.~4, pp. 7469--7476,
  2021.

\bibitem{xu2021fast}
W.~Xu and F.~Zhang, ``Fast-lio: A fast, robust lidar-inertial odometry package
  by tightly-coupled iterated kalman filter,'' \emph{IEEE Robotics and
  Automation Letters}, vol.~6, no.~2, pp. 3317--3324, 2021.

\bibitem{geiger2013vision}
A.~Geiger, P.~Lenz, C.~Stiller, and R.~Urtasun, ``Vision meets robotics: The
  kitti dataset,'' \emph{The International Journal of Robotics Research},
  vol.~32, no.~11, pp. 1231--1237, 2013.

\bibitem{nguyen2022ntu}
T.-M. Nguyen, S.~Yuan, M.~Cao, Y.~Lyu, T.~H. Nguyen, and L.~Xie, ``Ntu viral: A
  visual-inertial-ranging-lidar dataset, from an aerial vehicle viewpoint,''
  \emph{The International Journal of Robotics Research}, vol.~41, no.~3, pp.
  270--280, 2022.

\bibitem{quigley2009ros}
M.~Quigley, K.~Conley, B.~Gerkey, J.~Faust, T.~Foote, J.~Leibs, R.~Wheeler,
  A.~Y. Ng \emph{et~al.}, ``Ros: an open-source robot operating system,'' in
  \emph{ICRA workshop on open source software}, vol.~3, no. 3.2.\hskip 1em plus
  0.5em minus 0.4em\relax Kobe, Japan, 2009, p.~5.

\bibitem{grupp2017evo}
M.~Grupp, ``evo: Python package for the evaluation of odometry and slam.''
  \url{https://github.com/MichaelGrupp/evo}, 2017.

\end{thebibliography}

\end{document}